\documentclass[fleqn,10pt]{wlscirep}
\usepackage[utf8]{inputenc}
\usepackage[T1]{fontenc}

\usepackage{array}
\usepackage{adjustbox}

\usepackage[framemethod=TikZ]{mdframed}

\newmdenv[backgroundcolor=gray!20, linecolor=black] 
{mybox}

\newmdenv[backgroundcolor=white!20, linecolor=white, leftmargin=-1cm, rightmargin=-1cm] 
{mybox2}

\usepackage{subcaption}

\urldef{\myurl}\url{https://bioportal.bioontology.org/ontologies/IDO/?p=classes&conceptid=http%3A%2F%2Fpurl.obolibrary.org%2Fobo%2FVSMO_}

\title{An Epidemiological Knowledge Graph extracted from the World Health Organization's Disease Outbreak News}

\author[1,*]{Sergio Consoli}
\author[1,2]{Pietro Coletti}
\author[1,3]{Peter V. Markov}
\author[1]{Lia Orfei}
\author[1]{Indaco Biazzo}
\author[1]{Lea Schuh}
\author[1]{Nicolas Stefanovitch}
\author[1]{Lorenzo Bertolini}
\author[1]{Mario Ceresa}
\author[1]{Nikolaos I. Stilianakis}

\affil[1]{European Commission, Joint Research Centre (JRC), Ispra, Italy}
\affil[2]{Universit\`e catholique de Louvain, Institute of Health and Society (IRSS), Brussels, Belgium}
\affil[3]{London School of Hygiene and Tropical Medicine (LSHTM), London, United Kingdom}

\affil[*]{corresponding author(s): Sergio Consoli (sergio.consoli@ec.europa.eu)}

\begin{abstract}
The rapid evolution of artificial intelligence (AI), together with the increased availability of social media 
and news for epidemiological surveillance, 
are marking a pivotal moment in epidemiology and public health research. 
Leveraging the power of generative AI, we use an ensemble approach which incorporates multiple 
Large Language Models (LLMs) to extract valuable actionable epidemiological information from the World Health Organization (WHO)
Disease Outbreak News (DONs). DONs is a collection of regular reports on global outbreaks curated by the WHO 
and the adopted decision-making processes 
to respond to them. 
The extracted information is made 
available in a daily-updated dataset and a knowledge graph, referred to as \emph{eKG}, derived to provide a nuanced representation of the public health domain knowledge. 
We provide an overview of this new dataset 
and describe the structure of eKG, along with the services and tools used to access and utilize the data that we are 
building on top.
These innovative data resources 
open altogether new opportunities for epidemiological research, and
the analysis and surveillance of disease outbreaks. 
\end{abstract}
\begin{document}

\flushbottom
\maketitle

\thispagestyle{empty}


\section*{Background \& Summary}

We are currently 
at a critical transformative time point in the evolution of epidemiology and public health research \cite{salathe2012}, marked by the advent of artificial intelligence (AI) tools \cite{Brownstein20231597}. This advancement 
promises to fundamentally reshape the landscape of epidemiological studies, the manner in which infectious disease outbreaks are tracked, and our response to them. Yet, processing the vast quantity of unstructured epidemiology datasets 
poses significant challenges \cite{polgreen2008,Warsame2020}. 
In this paper we focus on the World Health Organization (WHO) Disease Outbreak News (DONs),
an official record of outbreaks maintained by the WHO and operational since 1996 \cite{WHO2024}. 
This publicly accessible system disseminates information on events related to health emergencies, 
reported by national authorities and various collaborators \cite{Oppenheim2019,Mondor20121184}. The WHO compiles these data into narrative descriptions that detail the specifics of each outbreak. Typically, these narratives pinpoint the affected geographical area, the disease in question or a syndromic occurrence with an undetermined cause, and may include details on the response efforts, such as the status of laboratory testing for confirmation. 
Despite their 
considerable information value, DONs were underutilized in academic research, primarily due to their unstructured, prose-based format, which poses significant barriers to systematic analysis and integration into automated health informatics systems \cite{LugoRobles3}.


In response to these challenges, our study employs an ensemble of advanced generative AI techniques, leveraging the strengths of multiple Large Language Models (LLMs) \cite{Vaswani20175999,Brown2020}, to extract and structure critical epidemiological information from the WHO DONs \cite{Consoli2024241ICICT}. 
The epidemiological information (i.e., disease and pathogen name, involved country, date of event, case totals, and mortality) extracted via this novel approach 
are made publicly available through a dataset updated daily, adopting \emph{FAIR} (\emph{F}indable, \emph{A}ccessible, \emph{I}nteroperable, and \emph{R}eusable) principles and following the paradigms of Linked Open Data (LOD) \cite{Heath20111}. 

We adopt a knowledge-centric paradigm \cite{10.1145/3227609.3227689,osborne2023kg}, involving the creation of a structured, interconnected, and formal representation of the WHO DONs to improve the capabilities of conducting complex and extensive analyses of disease outbreaks. 
Knowledge graphs (KGs) are data structures that describe the key entities in a domain and their relationships, presenting information in a format accessible and interpretable by both machines and humans \cite{Hogan2021}. The relationship between two entities is typically formalized as a triple in the format of \emph{<subject, predicate, object>} following the Resource Description Framework (RDF) \cite{RDF2014} 
as a standard model \cite{Ji2022494} (e.g., \emph{<Chikungunya virus, cause, 109 deaths>} or \emph{<Yersinia Pestis, occur, China>}) 
offering significant support to AI systems across multiple domains \cite{Tiwari20231}. 
Following this approach, an epidemiological knowledge graph (\emph{eKG}) is derived from DONs offering a case in point for a nuanced representation of the public health domain of knowledge, and providing a structured, interconnected, and formalized framework 
integrating the extracted epidemiological information from 
DONs. 

\begin{figure}[!b]
\centering
\includegraphics[width=\linewidth]{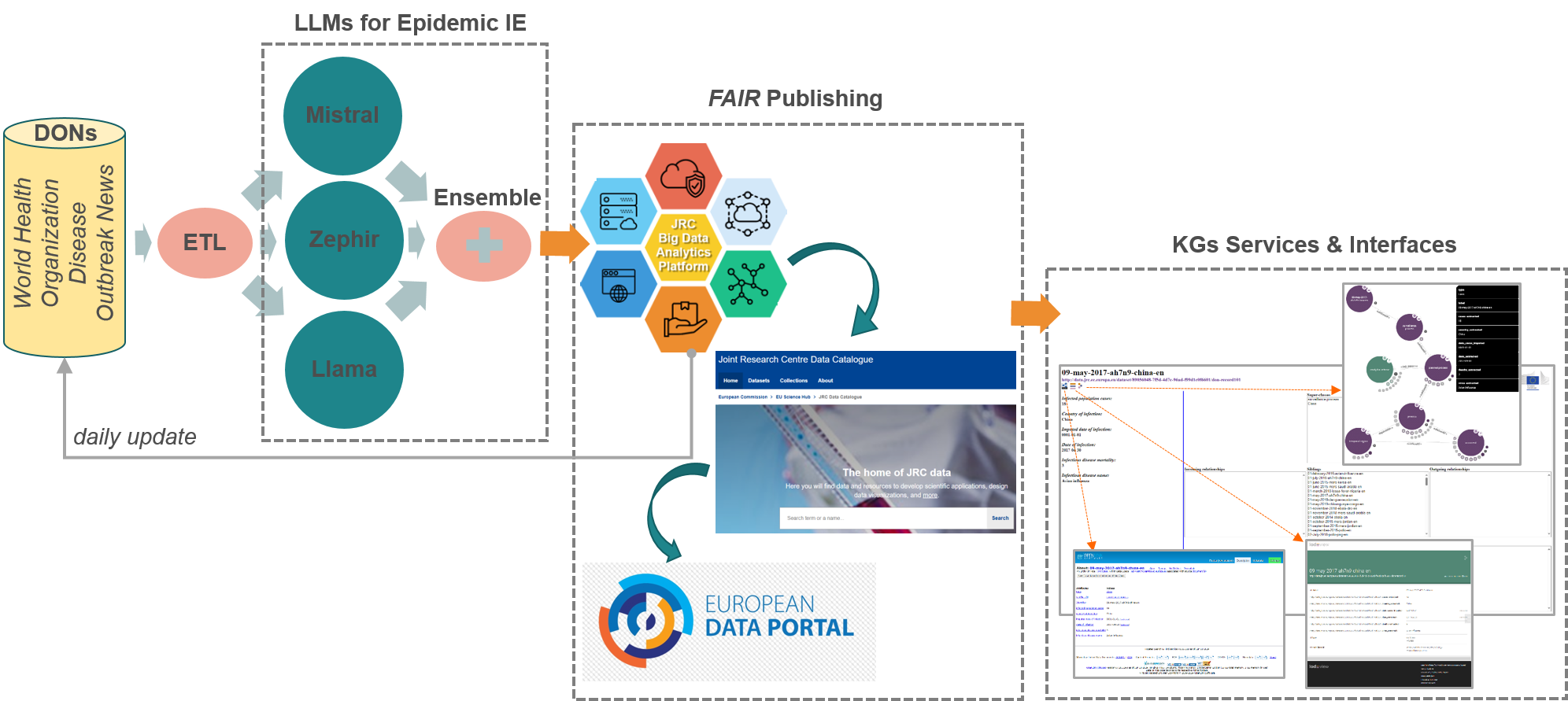}
\caption{Schematic overview of the pipeline to extract and publish critical epidemiological information from the WHO's Disease Outbreak News.}
\label{fig:SchematicOverview}
\end{figure}

Figure \ref{fig:SchematicOverview} provides a schematic overview of the pipeline, which will be described in more detail in the ``Methods'' section. A daily process is triggered to extract, transform and load (ETL) new reports from WHO DONs, which are then processed by the ensemble of LLMs for the epidemiological information extraction (IE) task. The extracted information are then published using FAIR principles along with the derived eKG. This enables a number of semantic services and interfaces to be deployed on top, allowing access and exploration of the derived knowledge graph. 
These resources have significant potentials to improve the utility and usability of WHO DONs for both researchers and policymakers, and promote new research opportunities in public health and epidemiology for the analysis of disease outbreaks, surveillance, and response to public health threats. 

The broader goals that motivate the creation of this knowledge graph include supporting the rapid identification of emerging health threats, enabling more effective international collaborations, and accelerating the development of public health interventions. By making the data available in a structured format, we aim to contribute to the global effort to prevent outbreaks from becoming epidemics or pandemics. This is particularly important in the early stage of an outbreak, as also demonstrated by the recent Covid-19 pandemic \cite{McDonald2021,pepe2020}, when timely and accurate information is critical to policy-makers for mobilizing resources, coordinating responses, and preparing the public for potential risks \cite{kissler2020}.

In summary, this data article makes a contribution in the field of epidemiology by providing a comprehensive and structured dataset derived from the WHO DONs. Our work further serves as a case in point for the value of extracting precious novel quantitative epidemiological data from the overabundant repositories of narrative reports. The establishment of a knowledge graph of epidemiological information and the application of AI to process and analyze unstructured data mark a transformative step towards enhancing our global understanding and management of disease outbreaks.

\section*{Methods}
%

The WHO Disease Outbreak News (DONs) platform is a critical source of information on confirmed acute public health events or potential events of concern. Since its launch in January 1996, the publicly accessible platform provided timely updates on various public health incidents. The core focus of DONs is to share news about confirmed or potential public health events, including unexplained incidents with significant international health implications, diseases with known causes that have the potential to spread globally, and events that could disrupt public health interventions, international travel, or trade. With over 3,000 curated epidemic news articles published since its inception, DONs established itself as a reliable and authoritative source of information on public health events.

To facilitate global health research that utilizes and analyzes DONs as a comprehensive record of outbreaks and a valuable repository of documents, we developed a standardized framework for sharing epidemiological
information, such that to allow DONs to be more useful to researchers and policymakers. The released resource was built using the automatic pipeline depicted in Figure \ref{fig:SchematicOverview}, composed of several modules that can be grouped into three main categories: \textit{i) LLMs for Epidemiological Information Extraction (IE)}, where the important epidemiological information is extracted from DONs using an ensemble of open-source LLMs, \textit{ii) FAIR Publishing}, which converts the extracted information into the eKG knowledge graph following the paradigm of Linked Open Data, adds additional contextual information to the extractions, and makes the resource available to local authorities, research organizations, national and European governmental institutions and international organizations through a public repository, and \textit{iii) KGs Services \& Interfaces}, consisting of several additional services and tools that we are currently building on top to expand the final user experience.

%

\subsection*{LLMs for Epidemic IE}

LLMs are a class of generative AI models that employ deep learning architectures, characterized by billions of parameters, to predict the likelihood of word sequences. These predictions are informed by the model's prior learning from extensive text corpora. The development of LLMs was significantly advanced by the introduction of the Transformer architecture \cite{Vaswani20175999}, a type of model design used in machine learning and AI. This architecture is particularly well-suited for handling sequential data, thanks to its attention mechanism \cite{sutskever2014sequence}, which enables the model to selectively focus on different parts of the input data. 
Unlike traditional sequential models like Recurrent Neural Networks (RNNs) and Long Short-Term Memory (LSTMs), Transformers can be trained in a highly parallelized and efficient manner, while also being particularly effective at capturing long-range dependencies in sequences, making them a powerful tool for natural language processing tasks \cite{Vaswani20175999,sutskever2014sequence}.

Building on the assessment of recent LLMs for epidemiological information extraction in \cite{Consoli2024241ICICT}, our work employs an ensemble approach, which was shown \cite{Consoli2024241ICICT} to be comparable or even superior to commercial techniques 
for extracting structured epidemiological information from text, including outbreak disease names, involved countries, event dates, total cases, and deaths. 
A solution based on open-source LLMs is ideal in this context because it overcomes specific usage constraints of commercial models, like, e.g. maximum number of tokens that can be processed per day, maximum rate of API calls allowed per minute etc., allowing for a full deployment on the entire DONs data.

By leveraging the extracted information in the context of infectious disease outbreaks from DONs, and integrating it into a surveillance system, we aim to improve the accuracy and timeliness of epidemiological modeling. 
A daily process is triggered to extract, transform and load (ETL) new reports from WHO DONs, opportunely cleaned and normalized, which are then processed by the ensemble of those LLMs. The adopted LLMs for the ensemble are: \emph{Mistral-7B-OpenOrca}, \emph{Zephyr-7B-Beta}, and \emph{Meta-Llama-3-70B-Instruct}. Please note that the original LLMs adopted for the ensemble in \cite{Consoli2024241ICICT} were: \emph{Mistral-7B-OpenOrca}, \emph{Zephyr-7B-Alpha}, and \emph{Meta-Llama-2-70B-Chat}. Here we adopt the upgraded, larger versions of these models. In the following sections we provide a brief description of these models.

\subsubsection*{Mistral-7B-OpenOrca} \label{Mistral-7b-open} 

The \emph{Mistral-7B-OpenOrca} model \cite{lian2023mistralorca1}, 
an open-source variant fine-tuned by OpenOrca on its proprietary datasets \cite{OpenOrca}, 
is built on top of the decoder-only Mistral LLM \cite{mukherjee2023orca}. Mistral models make use of a significant number of computational advancements with respect to other state-of-the-art LLMs, e.g. sliding-windows attention \cite{Beltagy2020Longformer}, which allows the model to dramatically extend the number of tokens it can simultaneously process. Despite having a relatively smaller size of 7 billion parameters, \emph{Mistral-7B-OpenOrca} demonstrates impressive performance across a range of tasks, while also providing fast inference speeds. Notably, its performance is often comparable to larger models, including the Llama 3 models family, and even rivals some models in the 20-30 billion parameter range \cite{Jiang2023Mistral7}. With a context length of 4,096 tokens, the \emph{Mistral-7B-OpenOrca} model stands out as one of the top-performing open-source models in its class, making it an attractive choice for various API applications due to its optimal balance of size, speed, and performance.

\subsubsection*{Zephyr-7B-Beta} \label{Zephyr-7B-Beta} 

The \emph{Zephyr-7B-Beta} model is an open-source LLM 
leveraging the transformer architecture and fine-tuned on a combination of publicly available and synthetic datasets on top of the Mistral LLM \cite{tunstall2023zephyrdirectdistillationlm}. This model employs a unique training approach, utilizing a combination of masked language modeling and next sentence prediction objectives, allowing it to learn a rich representation of language and generate coherent and context-specific text. Similarly to the \emph{Mistral-7B-OpenOrca} model, this model has also 7 billion parameters, 
which are distributed across 24 layers of transformer blocks, each consisting of self-attention mechanisms and feed-forward neural networks. Notably, this model utilizes a variant of the attention mechanism known as scaled dot-product attention \cite{tunstall2023zephyrdirectdistillationlm}, which enables it to efficiently process long-range dependencies in input sequences. Although being compact in size, it often outperforms larger LLMs in various tasks, demonstrating performance comparable to several models in the 20-30B range. The model is quick at inference and can be used in various types of tasks, making it an attractive option for deployment in resource-constrained environments. With a context length of 4,096 tokens, the \emph{Zephyr-7B-Beta} model ranks among the best open-source models available in terms of performance for models under 30 billion parameters, making it an efficient LLM for many API use-cases due to its size, speed, and efficiency.

\subsubsection*{Meta-Llama-3-70B-Instruct} \label{Meta-Llama-3-70B-Instruct} 

The \emph{Meta-Llama-3-70B-Instruct} model \cite{dubey2024llama3herdmodels} is a state-of-the-art open-source LLM developed by Meta, 
leveraging the transformer architecture and building upon the success of its predecessors \cite{touvron2023llama}. It is one of the most powerful open LLMs currently available, competing the level of performance of the highly acclaimed GPT models family. The base Llama model was pretrained on a diverse range of publicly available online data sources. This vast and varied data foundation equips the model with a comprehensive understanding of language structures and contexts. The fine-tuned \emph{Meta-Llama-3-70B-Instruct} model further enhances this understanding by leveraging publicly available instruction datasets and over 1 million human annotations. This model is fine-tuned on this massive dataset of instructions, which enables it to excel in tasks that require precise and context-specific responses \cite{llama3modelcard}. 
The fine-tuning process involves a combination of masked language modeling and next sentence prediction objectives, allowing the model to learn a rich representation of language and generate coherent and context-specific text, ensuring a high level of precision and adaptability in its responses.\\
The \emph{Meta-Llama-3-70B-Instruct} model boasts an impressive 70 billion parameters, which are distributed across 32 layers of transformer blocks, each consisting of self-attention mechanisms and feed-forward neural networks. With a context length of 4,096 tokens, this model is capable of handling complex language tasks, making it an ideal choice for applications that require advanced language understanding and generation capabilities. Furthermore, the model's instruction-based fine-tuning process ensures that it is highly adaptable and can respond accurately to a diverse range of prompts and queries, solidifying its position as a top-tier LLM in the field and making it a powerful resource for developing sophisticated AI applications.


\subsubsection*{Ensemble} \label{ensemble} 

Ensemble learning is an AI paradigm where multiple models are 
used to solve the same problem. In contrast to ordinary machine learning approaches which try to learn one hypothesis from training data, ensemble methods attempt to construct a set of hypotheses and combine them to use \cite{Sagi2018}. The \emph{Ensemble} model is a sophisticated AI approach that leverages the power of the multiple high-performing open-source LLMs, i.e. \emph{Mistral-7B-OpenOrca}, \emph{Meta-Llama-3-70B-Instruct}, and \emph{Zephyr-7B-Beta}, to generate the final outputs. Practically it uses a majority voting system to determine the best result from the contributed models \cite{Zhou20121}. 
Each of these models bring unique strengths and capabilities to \emph{Ensemble}, enhancing its overall performance. \emph{Mistral-7B-OpenOrca} excels in a range of tasks and has a reputation for its robustness and versatility. \emph{Zephyr-7B-Beta}, on the other hand, is known for its advanced capabilities and high-level performance. Lastly, \emph{Meta-Llama-3-70B-Instruct} is renowned for its efficiency in chat-based applications, providing high-quality, contextually accurate responses, further bolstering the ensemble's effectiveness.

%
%

\noindent
Each LLM is separately launched daily upon the new epidemiological DONs reports.
All LLMs share the same context length of 8K tokens, meaning they can process at one time a text of maximum length of 8K tokens when generating a response or performing a task. To note, the number of tokens corresponding to a certain number of words can vary greatly depending on the language, the tokenization method used, and the specific text. In English, with common tokenization methods, 100 tokens corresponds to approximately 75 words. 
Therefore, as a rough estimate, 8K tokens would correspond to around 6K words. 
%
In the case the DON report to process exceeds this context length value, it is split in pieces which then undergo a preliminary summarization step, using the following prompt: 

\begin{center}
\begin{tabular}{|@{\hspace{5mm}}>{\itshape}p{0.80\textwidth}@{\hspace{5mm}}|} 
\hline
\textbf{Prompt:} Summarize the epidemiological text below, focusing on any aspects that are relevant to the disease outbreak, place and time of the infectious disease outbreak occurred, 
and the number of cases and deaths derived. \\
Do not invent. 
Write no explanations or notes . \\
\hline
Text: ... \\
\hline
\end{tabular}
\end{center}


\noindent
Afterwards, for the extraction of the key epidemiological entities (name of disease outbreaks, involved country, date of the outbreak, number of confirmed cases and deaths) 
for each DON report, the LLMs are supplied with the prompt depicted below: 

\begin{center}
\begin{tabular}{|@{\hspace{5mm}}>{\itshape}p{0.80\textwidth}@{\hspace{5mm}}|} 

\hline
\textbf{Prompt:} From the text below extract the following items:\\
1 - The name of the disease that caused the outbreak. \\
2 - The name of the country where this disease outbreak occurred, if present. \\
3 - The date when this disease outbreak occurred, if present. Show the date in the format YYYY-mm-dd. \\
4 
- The number of deaths caused exclusively by the disease outbreak mentioned in the text, if present. \\
Format your response as a JSON object with the following keys: disease name, country, date, cases. 
If the information is not present, do not invent and use ``None'' as the value. \\
\hline
Text: ... \\
\hline
\end{tabular}
\end{center}

\noindent

As an example, consider the WHO DON in \href{https://www.who.int/emergencies/disease-outbreak-news/item/31-may-2018-nipah-virus-india-en}{Nipah virus - India}, reporting a Nipah virus outbreak in Kerala, India, resulting in 15 laboratory-confirmed cases, 13 deaths and 2 hospitalizations, with bats linked to the initial transmission, prompting a robust local and WHO-supported public health response, including surveillance and infection control measures. 
The extraction answer provided in JSON by the \emph{Mistral-7B-OpenOrca} model using the detailed prompt was:
\begin{center}
\{ \textit{``disease'': ``Nipah virus'', ``country'': ``India'', ``date``: ``2018-05-19'', ``cases'': ``15'', ``deaths'': ``13''} \},
\end{center}
which we can note correctly structures the main epidemiological information occurred in this disease outbreak. 

\noindent
After a report is processed by all the contributed LLMs, majority voting is used among their outputs for each of the extracted epidemiological information, forming in this way the output of the \emph{Ensemble}. 
Ensemble methods combining the predictions from multiple models are able to improve the overall performance. This is indeed a well-documented phenomenon in machine learning \cite{Sagi2018,Zhou20121}. The underlying principle is that by pooling together the strengths and mitigating the weaknesses of various models, the ensemble can achieve better accuracy and robustness than any single model alone. 
%
%
For example, suppose the output for the extracted number of cases from the three LLMs for a report would be: 15, 17, and 15, then \emph{Ensemble} would give 15 as its output for the number of cases for that report. When textual epidemiological information, that is disease names and countries, need to be evaluated by the majority voting system, the process is slightly more complicated, as two terms representing semantically the same concept should be grouped together. This was achieved by compiling a dictionary of synonyms of disease and pathogen names, and another of country names synonyms. Synonyms are created by (i) first checking for syntactic similarity among the terms to be compared (e.g. ``Trinidad and Tobago'' and ''Trinidad \& Tobago``, or ''Florida, USA`` and ''USA (Florida)``; (ii) checking for synonymy by looking for overlapping synsets in WordNet \cite{Miller199539}; 
(iii) checking for semantic similarity among the terms to be compared (e.g. ''Middle East respiratory syndrome`` and ''MERS-CoV``, or ''Dengue`` and ''Breakbone fever``. In order to calculate the semantic similarity we used Sentence Transformers, also known as SBERT (Sentence-BERT) \cite{reimers-2019-sentence-bert}, 
which is a modification of the pre-trained BERT (Bidirectional Encoder Representations from Transformers) \cite{devlin-etal-2019-bert} network that uses siamese and triplet network structures to derive semantically meaningful sentence embeddings. These embeddings can then be compared using a metric like the cosine similarity \cite{Korst201939} to determine the semantic similarity between sentences. 
SBERT fine-tunes the BERT network on a dataset of sentence pairs, training it to produce embeddings that bring semantically similar sentences closer together in the embedding space. This allows for much faster performance on downstream tasks, as the sentence embeddings can be precomputed and easily compared without the need for further BERT processing. In particular, to evaluate the semantic similarity among country names, we experimentally selected the \emph{all-mpnet-base-v2} \cite{mpnetbase2024} 
SBERT model, which is an all-round model trained on a large and diverse dataset of over 1 billion training pairs and tuned for many diverse use-cases. For the disease outbreaks we used instead BioBERT \cite{BioBERT-Lee20201234}, a domain-specific version of the BERT model pre-trained on biomedical literature to enhance performance on biomedical natural language processing tasks \cite{deka2022evidence}. 
We set an experimental threshold of $0.8$ for the semantic similarity $\in[0,1]$, meaning that two terms having a semantic similarity larger than this threshold were considered to represent the same concept.
In this way, clusters of disease names and countries synonyms are obtained, forming dictionaries of similar concepts.

\noindent
This \emph{Ensemble} model, with its majority voting mechanism, ensures that the most agreed-upon output from the three LLMs is chosen, thereby increasing the likelihood of accuracy and reliability in its predictions. 
In \cite{Consoli2024241ICICT}, the authors proved that this approach improves upon single open-source LLMs for epidemiological IE and is able to compete with the performance of more advanced commercial LLMs.

%


\subsection*{\emph{FAIR} Publishing}

The epidemiological knowledge graph, eKG, was produced from the information 
extracted by the WHO DONs, providing a structured, interconnected, and formalized data model for the public health domain. It was implemented by using Semantic Web technologies, such as RDF and the Web Ontology Language (OWL) \cite{Antoniou2004OWL}, which allow human experts to verify, curate, and correct both the data and their 
schema.
For the eKG extraction, we strongly follow principles from ontology design patterns (ODPs) \cite{Gangemi2005262} and LOD \cite{Heath20111}, 
that is, whenever possible we reuse existing ODPs from online LOD repositories. Reused patterns are annotated with the OPLa framework \cite{Shimizu201823,Asprino2021} in order to facilitate class mapping and identification. 
In particular we reuse classes and properties from IDO (Infectious Disease Ontology) \cite{Cowell2010373IDO} 
and GeoNames \cite{GeoNames2024}, 
and include explicit alignments to them by \texttt{rdfs:subClassOf} and \texttt{owl:sameAs} axioms. IDO is a collection of structured entities generally relevant to both the biomedical and clinical aspects of infectious diseases. 
The structure of IDO adheres to the Basic Formal Ontology (BFO) \cite{BFO} 
and aims to provide a coverage of the infectious disease domain. Terms in IDO that are within the scope of other OBO Foundry ontologies \cite{Jackson2021OboFoundry} 
are derived from those respective ontologies. 
In this way we also map indirectly to the WHO's International Classification of Diseases, Version 10 (ICD10) \cite{ICD10}. 
GeoNames is instead an open-source geographical database, updated regularly, covering all countries worldwide and containing over eleven million placenames, that we use to describe geographical entities, such as cities, regions, or countries names where a disease outbreak has place \cite{Krauer2022GeoNamesEpi}. 
An in-depth explanation of the structure of eKG and each associated data record is given in the ``Data Records'' section, 
providing also an overview of the data files and their formats.

\begin{figure}[!b]
\centering
\includegraphics[width=1.0\linewidth]{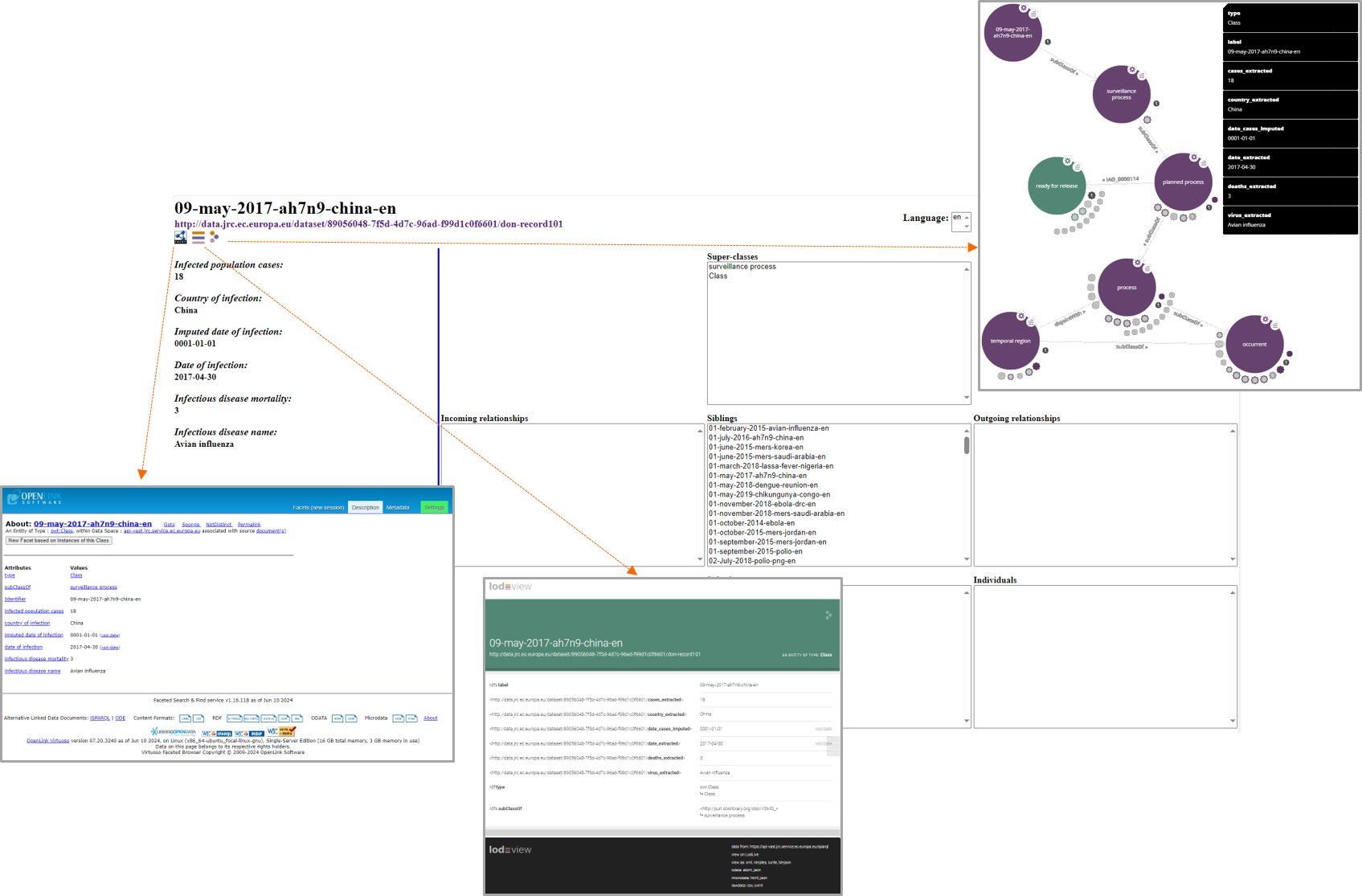}
\caption{Snippet of eKG exploration services and interfaces.}
\label{fig:kgservices}
\end{figure}

\noindent
The disease outbreak events are extracted daily (at night) from the new DONs reports by the LLMs ensemble via the JRC Big Data Analytics Platform \cite{Soille201830}, 
where they are also mapped as individuals of eKG. 
The epidemiological knowledge graph, currently accounting to around 2.9K disease outbreak events, is available at \cite{JRCdataEPIKG} in structured RDF/XML and Turtle \cite{RDFturtle2014} formats, with raw extractions also given in CSV, within the Joint Research Centre Data Catalogue \cite{JRCDataCatalogue2024} 
at the following permanent location:
\url{http://data.jrc.ec.europa.eu/dataset/89056048-7f5d-4d7c-96ad-f99d1c0f6601}, used also as reference namespace for the eKG individuals. The KG was also released within the European Data portal \cite{Kirstein2019192EuropeanDataPortal}. 
the official data repository for European data, at the following permanent location: \url{https://data.europa.eu/data/datasets/89056048-7f5d-4d7c-96ad-f99d1c0f6601}.

\subsection*{KGs Services \& Interfaces}

Using the described pipeline, we generated the knowledge graph of epidemiological information, comprising roughly of 26k distinct (non-reified) triples, representing around 2.9k unique outbreak events via a total of 2.3 generalized axioms. 
To access the data programmatically, we set up a SPARQL \cite{SPARQLPerez2009} 
endpoint, available at: \url{https://api-vast.jrc.service.ec.europa.eu/sparql/} \cite{sparqlEndpoitEPIKG}, 
where eKG can be queried and analytical information on target disease outbreaks, attributes, and relations can be retrieved in structured data format, as better described in the ``Usage Notes'' section. On top of eKG we are currently developing a number of intuitive and user-friendly services and interfaces at European Commission premises, 
including content negotiation, visualization and navigation of data, improved exploitation of the available information and knowledge discovery. 
%
The central point of access (Figure \ref{fig:kgservices}) is a web interface where it is possible to have an immediate view of the main epidemiological information associated to a disease outbreak, along with quick reference pointers to other semantic applications, namely: the \emph{Virtuoso Faceted Browser}, available at \url{https://api-vast.jrc.service.ec.europa.eu/fct/}, 
a general purpose RDF data query facility service for faceted browsing over entity relationship types, 
allowing users to explore 
the outbreak events in a structured and intuitive way; 
\emph{LodView} \cite{Kirillovich2022143LodView}, 
a web application compliant to World Wide Web Consortium (W3C) \cite{Griset2011353W3C} standards for IRI dereferentiation which provides a representation of our RDF disease outbreak resource through a custom intuitive HTML page. The tool implements content negotiation of eKG resources, and eventually allows to download the selected 
disease oubreak in various needed formats, such as, e.g., \textit{xml}, \textit{ntriples}, \textit{turtle}, and \textit{ld+json}; \emph{LodLive} \cite{Camarda2012197LodLive}, 
a navigator of RDF resources based on a graph layout. It is used for connecting RDF browser capabilities with the effectiveness of data graph representation, 
enabling the navigation of the eKG resources starting from a specific disease outbreak event. 
Examples of use of these tools are given in the ``Usage Notes'' section.

\section*{Data Records}

The extracted epidemiological information from DONS and the knowledge graph are available respectively in CSV, RDF/XML and Turtle formats \cite{JRCdataEPIKG} within the Joint Research Centre Data Catalogue \cite{JRCDataCatalogue2024}. 
We enable 
the interested community 
to access and use the produced data and ontology under Creative Commons Attribution 4.0 International (CC BY 4.0) license \cite{CCBY4}. 

\begin{center}
\begin{tabular}{|@{\hspace{5mm}}>{\itshape}p{0.90\textwidth}@{\hspace{5mm}}|} 
\hline
\vspace*{0.1cm}
Repository name: Joint Research Centre Data Catalogue\\
Data identification number:\\
\url{http://data.jrc.ec.europa.eu/dataset/89056048-7f5d-4d7c-96ad-f99d1c0f6601}\\
\vskip 0.1cm
Repository name: European Data portal\\
Data identification number:\\
\url{https://data.europa.eu/data/datasets/89056048-7f5d-4d7c-96ad-f99d1c0f6601}\\
\vskip 0.1cm
Digital Object Identifier (DOI):\\
\url{https://doi.org/10.2905/89056048-7f5d-4d7c-96ad-f99d1c0f6601}\\[0.3cm] 
\hline
\end{tabular}
\end{center}


\noindent
Data are freely available to the public and can be downloaded without any login details by selecting the appropriate URL in one of the repositories above. In particular, for the sake of clarity, we specify below the URL of each of the released data source, with the relative format in parenthesis:

\begin{itemize}
 \item Epidemiological IE from WHO DONs using LLMs (CSV): \url{https://jeodpp.jrc.ec.europa.eu/ftp/jrc-opendata/ETOHA/corpus_processed/SUMMARIZED/OutputAnnotatedTexts-LLMs-ENSEMBLE_whoDons.csv}
 \item Epidemiological knowledge graph (eKG) from WHO DONs using LLMs (RDF/XML): \url{https://jeodpp.jrc.ec.europa.eu/ftp/jrc-opendata/ETOHA/ETOHA-OPEN/epidemicIE-DONs.rdf}
 \item Epidemiological knowledge graph (eKG) from WHO DONs using LLMs - no imports (RDF/XML): \url{https://jeodpp.jrc.ec.europa.eu/ftp/jrc-opendata/ETOHA/ETOHA-OPEN/epidemicIE-DONs-Skeleton.rdf}
 \item Epidemiological knowledge graph (eKG) from WHO DONs using LLMs (Turtle): \url{https://jeodpp.jrc.ec.europa.eu/ftp/jrc-opendata/ETOHA/ETOHA-OPEN/epidemicIE-DONs.ttl}
\end{itemize}

\noindent
The raw CSV file contains the extracted details of the occurring event from each report. In particular, it contains the following information:
\begin{itemize}
\item ``fileid'' column, presents the ID of the report, which is taken from the slug of the URL of the DON item;
\item ``virus\_extracted'' column, denotes the extracted name of the disease outbreak extracted from the current DON item;
\item ``country\_extracted'' column, depicts the involved countries as extracted from the report linked to the specific disease outbreak; 
\item ``date\_extracted'' column, presents the date when the disease outbreak occurred, in the format YYYY/MM/DD; 
\item ``date\_cases\_Imputed'' column, shows the date reported in the fileid of the report, if present, and it can be related to the publication date of the report (YYYY/MM/DD format), although not formally defined in DONs. Please note that this information could be useful for analysis only when the ``date\_extracted'' is missing, that means that it was not possible to automatically extract from the main text of the report an actual date of the disease outbreak;
\item ``cases\_extracted'' column, shows the extracted number of confirmed cases deriving from the specific disease outbreak; 
\item ``deaths\_extracted'' column, presents the related event mortality. 
\end{itemize}

\begin{figure}[!ht]
\centering
\includegraphics[width=0.8\linewidth]{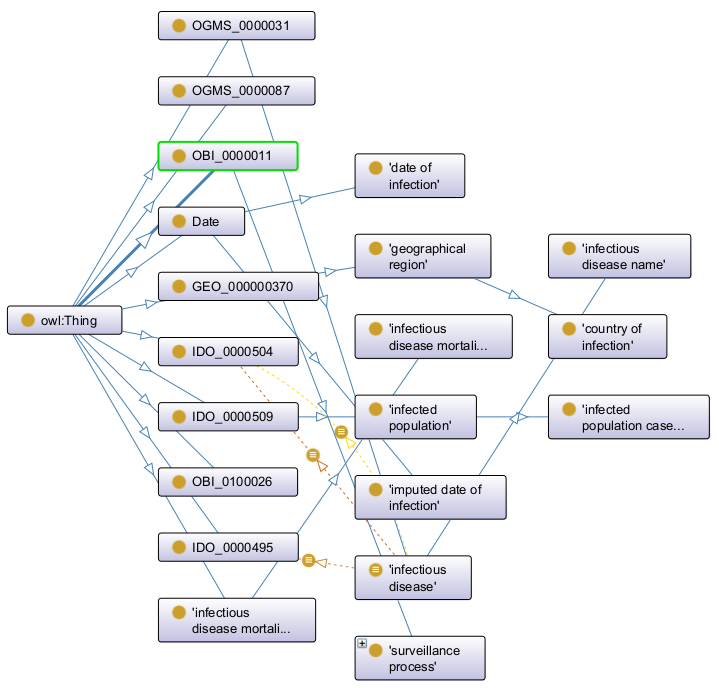}
\caption{Main classes of the epidemiological knowledge graph.}
\label{fig:ontograph}
\end{figure}

\noindent
In our data engineering work to model the knowledge graph of epidemiological information, we adhered to standard style guidelines for identifying and describing linked data resources \cite{gangemi2009ontology}. Specifically, within eKG, data names were expressed in lowercase and any potential space characters replaced with dashes, as per the established norms for naming and labeling knowledge graphs. In contrast, class names within ontology definitions were expressed in uppercase, following the common style guidelines used for such definitions.
The main classes of the knowledge graph are mapped to RDF/OWL as 
illustrated in Figure \ref{fig:ontograph}. A disease outbreak event from DONs is an \texttt{owl:Class} of type \texttt{IDO:surveillance\_process}, which is a specialization of a generic \texttt{OBI\_0000011:planned\_process}. A disease outbreak is associated with an \texttt{IDO\_0000436:infectious\_disease}, having a specific \texttt{infectious\_disease\_name} corresponding to the name of the disease extracted from the DONs text, occurring in a \texttt{GEO\_000000372:geographical\_region} in a \texttt{date\_of \_infection}, i.e. a \texttt{dcterms:date} \cite{Weibel2000DublinCore}. 
Please note that when a date is reported in the title of a DONs, such e.g. in the 
\href{https://www.who.int/emergencies/disease-outbreak-news/item/31-may-2018-nipah-virus-india-en}{Nipah virus - India} DON, 
an additional \texttt{imputed\_date\_of\_infection} is also associated to the disease outbreak event; however this information should be taken with the due care, given the reporting date might be delayed with respect to the actual occurring date of the disease outbreak. In the \href{https://www.who.int/emergencies/disease-outbreak-news/item/31-may-2018-nipah-virus-india-en}{Nipah virus - India} example, 
indeed, the reporting \texttt{imputed\_date\_of\_infection} would be the ``2018-05-31'', but the actual \texttt{date\_of\_infection} reported in the DON is the ``2018-05-19''. The \texttt{imputed\_date\_of\_infection} could be useful for analysis only when it was not possible to automatically extract from the main text of the report an actual \texttt{date\_of\_infection}. A disease outbreak event is then associated with a portion of \texttt{infected\_population} expressed in terms of \texttt{IDO\_0000511:infected \_population\_cases} extracted from the DON report, with can ultimately brings to an \texttt{IDO\_0000489:infectious \_disease\_mortality}, i.e. the extracted number of deaths from the disease outbreak report. 

\noindent
Consider the following vocabulary prefixes, in RDF/XML:
\begin{mybox2}
\texttt{<rdf:RDF\\
    \hspace*{8mm}xmlns=``http://data.jrc.ec.europa.eu/dataset/89056048-7f5d-4d7c-96ad-99d1c0f6601/''\\
    \hspace*{8mm}xml:base=``http://data.jrc.ec.europa.eu/dataset/89056048-7f5d-4d7c-96ad-f99d1c0f6601/''\\
     \hspace*{8mm}xmlns:owl=``http://www.w3.org/2002/07/owl\#''\\
     \hspace*{8mm}xmlns:rdf=``http://www.w3.org/1999/02/22-rdf-syntax-ns\#''\\
     \hspace*{8mm}xmlns:xml=``http://www.w3.org/XML/1998/namespace''\\
     \hspace*{8mm}xmlns:rdfs=``http://www.w3.org/2000/01/rdf-schema\#''\\
     \hspace*{8mm}xmlns:xsd=``http://www.w3.org/2001/XMLSchema\#''\\
     \hspace*{8mm}xmlns:skos=``http://www.w3.org/2004/02/skos/core\#''\\    
     \hspace*{8mm}xmlns:obo=``http://purl.obolibrary.org/obo/''  \\  
     >}
\end{mybox2}

\noindent
The RDF/XML triples of this example are encapsulated by the following OWL statement:
\begin{mybox2}
\texttt{
<owl:Class rdf:about=``http://data.jrc.ec.europa.eu/dataset/89056048-7f5d-4d7c-96ad-\\[-0.75cm]    
    \begin{flushright}f99d1c0f6601/don-record2740''>\end{flushright}
     \vskip -0.25cm  
\hspace*{8mm}<rdfs:subClassOf rdf:resource=``http://purl.obolibrary.org/obo/VSMO\_''/>\\
\hspace*{8mm}<rdfs:label>31-may-2018-nipah-virus-india-en</rdfs:label>\\
\hspace*{8mm}<virus\_extracted>Nipah Virus</virus\_extracted>\\
\hspace*{8mm}<country\_extracted>India</country\_extracted>\\
\hspace*{8mm}<date\_extracted rdf:datatype=``http://www.w3.org/2001/XMLSchema\#date''>2018-05-19\\[-0.75cm]    
    \begin{flushright}</date\_extracted>\end{flushright}
     \vskip -0.25cm  
\hspace*{8mm}<date\_cases\_Imputed rdf:datatype=``http://www.w3.org/2001/XMLSchema\#date''>2018-05-31\\[-0.75cm]    
    \begin{flushright}</date\_cases\_Imputed>\end{flushright}
     \vskip -0.25cm  
\hspace*{8mm}<cases\_extracted>15</cases\_extracted>\\
\hspace*{8mm}<deaths\_extracted>13</deaths\_extracted>\\
</owl:Class>
}
\end{mybox2}

\noindent
where the raw extractions metadata are linked the ontology classes of eKG as shown below:
\begin{mybox2}
\texttt{
<owl:Class rdf:about=``http://data.jrc.ec.europa.eu/dataset/89056048-7f5d-4d7c-96ad-\\[-0.75cm]    
    \begin{flushright}f99d1c0f6601/virus\_extracted''>\end{flushright}
     \vskip -0.25cm  
    \hspace*{8mm}<rdfs:subClassOf rdf:resource=``http://purl.obolibrary.org/obo/IDO\_0000436''/>\\
    </owl:Class>\\[0.4cm]
<owl:Class rdf:about=``http://data.jrc.ec.europa.eu/dataset/89056048-7f5d-4d7c-96ad-\\[-0.75cm]    
    \begin{flushright}f99d1c0f6601/country\_extracted''>\end{flushright}
     \vskip -0.25cm  
    \hspace*{8mm}<rdfs:subClassOf rdf:resource=``http://purl.obolibrary.org/obo/GEO\_000000372''/>\\
</owl:Class>\\[0.4cm]    
<owl:Class rdf:about=``http://data.jrc.ec.europa.eu/dataset/89056048-7f5d-4d7c-96ad-\\[-0.75cm]    
    \begin{flushright}f99d1c0f6601/date\_extracted''>\end{flushright}
     \vskip -0.25cm  
    \hspace*{8mm}<rdfs:subClassOf rdf:resource=``http://purl.org/dc/terms/date''/>\\
</owl:Class>\\[0.4cm]
<owl:Class rdf:about=``http://data.jrc.ec.europa.eu/dataset/89056048-7f5d-4d7c-96ad-\\[-0.75cm]    
    \begin{flushright}f99d1c0f6601/date\_cases\_Imputed''>\end{flushright}
     \vskip -0.25cm  
    \hspace*{8mm}<rdfs:subClassOf rdf:resource=``http://purl.org/dc/terms/date''/>\\
</owl:Class>\\[0.4cm]
<owl:Class rdf:about=``http://data.jrc.ec.europa.eu/dataset/89056048-7f5d-4d7c-96ad-\\[-0.75cm]    
    \begin{flushright}f99d1c0f6601/cases\_extracted''>\end{flushright}
     \vskip -0.25cm  
        \hspace*{8mm}<rdfs:subClassOf rdf:resource=``http://purl.obolibrary.org/obo/IDO\_0000511''/>\\
    </owl:Class>\\[0.4cm]
<owl:Class rdf:about=``http://data.jrc.ec.europa.eu/dataset/89056048-7f5d-4d7c-96ad-\\[-0.75cm]    
    \begin{flushright}f99d1c0f6601/deaths\_extracted''>\end{flushright}
     \vskip -0.25cm  
    \hspace*{8mm}<rdfs:subClassOf rdf:resource=``http://purl.obolibrary.org/obo/IDO\_0000489''/>\\
</owl:Class> }
\end{mybox2}

\noindent
We enable direct access to the extracted knowledge graph data by means of custom SPARQL queries via the dedicated REST APIs SPARQL endpoint service \cite{sparqlEndpoitEPIKG}. 

\begin{table}[!b]
\centering
\begin{tabular}{|l|l|l|}
\hline
\textbf{Top 10 Disease Outbreak Names} & \textbf{Top 10 Countries} & \textbf{Top 10 Disease Outbreak Names - Country pairs} \\ \hline\hline
Avian influenza (570)    & China (243)             & { \bf Saudi Arabia - MERSCoV} (201)           \\ \hline
MERSCoV (296)            & Democratic Republic of Congo (226)             & China - Avian influenza (172)          \\ \hline
Ebola (211)              & Saudi Arabia (225)      & { \bf Democratic Republic of Congo - Ebola} (134)                    \\ \hline
Cholera (146)            & Egypt (155)             & The Netherlands - Avian influenza (108)\\ \hline
Yellow Fever (146)       & The Netherlands (142)   & Egypt - Avian influenza (101)          \\ \hline
SARS (78)                & United Republic of Tanzania (89) & { \bf China-SARS} (51)               \\ \hline
Hemorrhagic fever (68)   & Viet Nam (66)           & Viet Nam - Avian influenza (41)        \\ \hline
Dengue Fever (56)        & Angola (60)             & Democratic Republic of Congo - Hemorrhagic fever (31)         \\ \hline
Meningococcal (Group C) (54) & Cameroon (51)      & { \bf Guinea - Ebola }(30)             \\ \hline
H5N1 (51)                & Guinea Bissau (49)      & Cameroon - Yellow Fever (30)           \\ \hline
\end{tabular}
\caption{The top reported disease outbreak names, countries, and disease outbreak name-country pairs by total number of extracted events (number given in parentheses). Pairs of disease outbreak name-country in {\bf bold} are further illustrated in Figure \ref{fig_4cases}.
\label{tab_top10}
}
\end{table}

\subsection*{Data Statistics}

\begin{figure}[!t]
\centering
\adjustbox{cfbox=gray 1pt}{
\includegraphics[width=0.75\linewidth]{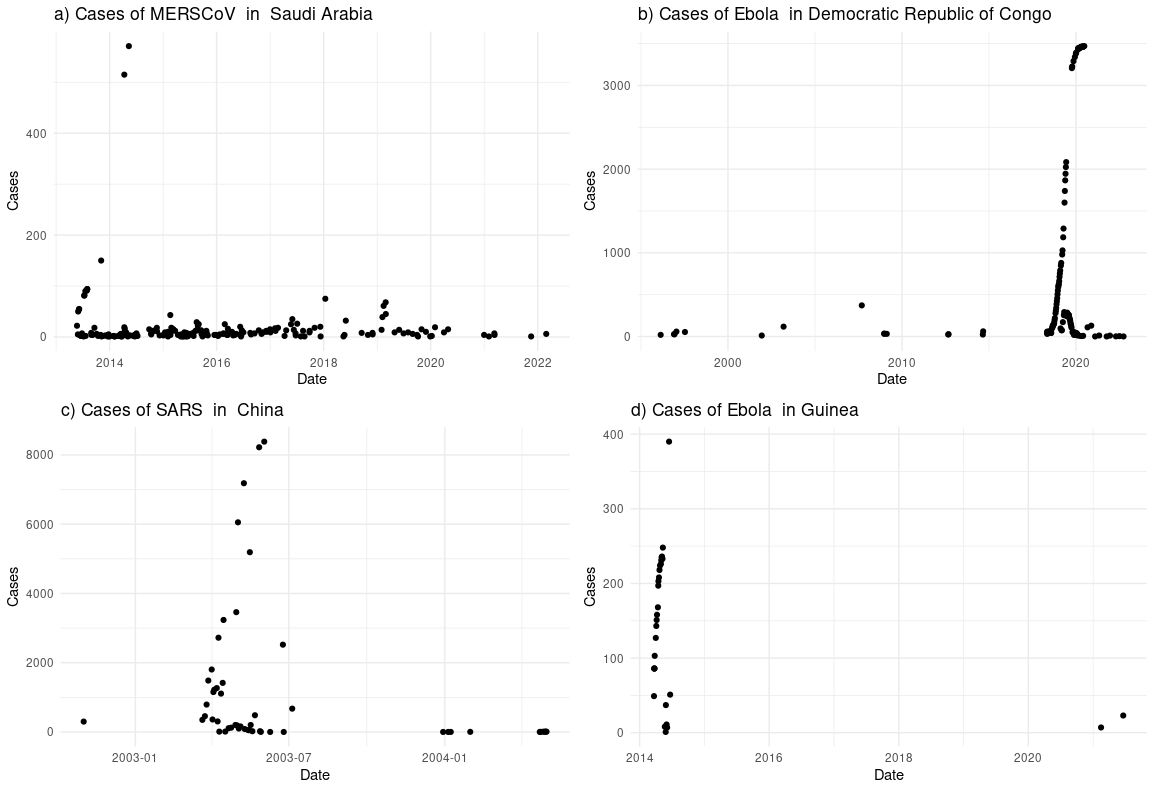}
}
\caption{Number of extracted cases vs. time for 4 case studies: MERS-Cov in Saudi Arabia (Panel a), Ebola cases in the Democratic Republic of Congo (Panel b), SARS cases in China (Panel c), and Ebola cases in Guinea (Panel d). These four case studies correspond to the Disease Outbreak Names - Country pairs reported in bold in Table \ref{tab_top10}.}
\label{fig_4cases}
\end{figure}

After removing entries with missing country and disease, the dataset consists of 2384 entries, listing 126 unique pathogens/diseases and 180 unique countries. Table \ref{tab_top10} shows the top 10 disease outbreak names, countries, and disease outbreak names - country pairs by total number of extracted events. The events detected contain a combination of human diseases and zoonoses, with worldwide coverage. Figure \ref{fig_4cases} presents four case studies as a time series of the number of cases over time. Figure \ref{fig_4cases}, Panel a, shows the numbers of MERS-Cov cases reconstructed by the LLM in Saudi Arabia, with a consistent reconstruction of the MERS-Cov cases experienced in the area \cite{WHO_mers} 
Panels b) and d) of Figure \ref{fig_4cases} show the number of Ebola cases reconstructed by the LLM in the Democratic Republic of the Congo (DRC) and Guinea, respectively, illustrating the Kivu Ebola Epidemic of 2018-2020 in DRC \cite{WHO_ebola_DRC_2018} 
and the West Africa outbreak of 2014-2016~\cite{WHO_ebola_Wafr_2014}. 
Panel c) of Figure \ref{fig_4cases} shows instead the reconstructed number of SARS cases in China, with a qualitative reconstruction of the SARS outbreak of 2002-2004~\cite{wikipedia_sars_2003}. Overall, the dataset is able to reconstruct the qualitative trends of several epidemics worldwide. For a more quantitative comparison, see the next ``Technical Validation'' section.

\section*{Technical Validation}
%

In this section, we present a formal evaluation of the quality of the extracted epidemiological data within eKG. 
%
%
While the generation process of the knowledge graph is entirely automated, 
it is important to assess 
the accuracy of the extraction methodology. This evaluation will serve as a support to the dissemination of eKG as a public resource. Building on the work done in \cite{Consoli2024241ICICT}, in this section we present a comparison of the performance of different LLMs for infectious disease information extraction in order to validate the adopted ensemble approach for producing eKG. The comparison was made over a subset of the Incident Database (IDB), a repository developed 
for the purpose of event-based surveillance \cite{Abbood2020}. 
The IDB is structured to house a comprehensive collection of data related to epidemic events, making it an important resource for researchers and healthcare professionals.
The database includes reports from various surveillance systems, including samples from WHO DONs, among others, 
that were meticulously annotated by domain experts. These annotations offer a valuable layer of insight, enabling researchers to access not only the raw data but also the expert interpretations of those data. Before being incorporated into the IDB, the data undergo a rigorous process of cleaning and standardization. This process ensures that all entries in the database conform to a standardized format, making the data easier to analyze and interpret. Additionally, data originally in foreign languages are translated to ensure accessibility to a broader audience. 
To evaluate the performance of the epidemiological information extraction models under investigation, a reference subset of the IDB, consisting of 171 samples for which we were able to reconstruct the source url and locate the main text of the report, was considered. This subset serves as a gold standard for benchmarking the performance of the LLMs.


\noindent
We differentiate the comparison for each of the epidemiological IE tasks, namely the extraction of the name of the disease, the confirmed case count, the country where the cases were reported, and the date of the case count. It was not possible to evaluate the task of extraction of the number of deaths as this information was not annotated in the IDB. Nonetheless, 
given the similarity of this task to the extraction of the number of cases, its performance is likely to be comparable.

\begin{table}[!b]
\centering
\begin{tabular}{|rl|r|r|r|}
\hline
 & & \multicolumn{1}{c|}{\textbf{Precision}} & \multicolumn{1}{c|}{\textbf{Recall}} & \multicolumn{1}{c|}{\textbf{F$_{1}$ score}} \\
\hline
\emph{GPT-3.5-Turbo-16k}	 & &	0.750	&	0.961	&	0.842 \\
\emph{GPT-4-32k}			 & &	0.756	&	0.945	&	0.840 \\
\emph{GPT-4-FewShots}		 & &	0.770	&	0.914	&	0.836 \\
\emph{Pythia-12B}			 & &	0.712	&	0.773	&	0.742 \\
\emph{MPT-30B-Chat}		 & &	0.740	&	0.891	&	0.809 \\
\emph{Meta-Llama-2-70B-Chat}	 & &	0.757	&	0.898	&	0.821 \\
\emph{Meta-Llama-3-70B-Instruct}	 & &	0.762	&	0.926	&	0.834 \\
\emph{Mistral-7B-OpenOrca}	 & &	0.755	&	0.891	&	0.817 \\
\emph{Zephyr-7B-Alpha}	 & &	0.752	&	0.875	&	0.809 \\
\emph{Zephyr-7B-Beta}	 & &	0.756	&	0.882	&	0.816 \\
\emph{Ensemble}		 & &	\textbf{0.794}	&	\textbf{0.988}	&	\textbf{0.851} \\
\hline
\end{tabular}
\caption{LLMs comparison for the information extraction task of the disease outbreak name over IDB.}\label{epidemicname}
\end{table}

\begin{table}[!b]
\centering
\begin{tabular}{|rl|r|r|r|}
\hline
 & & \multicolumn{1}{c|}{\textbf{Precision}} & \multicolumn{1}{c|}{\textbf{Recall}} & \multicolumn{1}{c|}{\textbf{F$_{1}$ score}} \\
\hline
\emph{GPT-3.5-Turbo-16k}	 &	&	0.981	&	0.910	&	0.944	\\
\emph{GPT-4-32k}			 &	&	0.981	&	0.928	&	0.954	\\
\emph{GPT-4-FewShots}		 &	&	0.981	&	0.928	&	0.954	\\
\emph{Pythia-12B}			 &	&	0.962	&	0.452	&	0.615	\\
\emph{MPT-30B-Chat}		 &	&	0.979	&	0.855	&	0.913	\\
\emph{Meta-Llama-2-70B-Chat}	 &	&	0.980	&	0.898	&	0.937	\\
\emph{Meta-Llama-3-70B-Instruct}	&	&	0.984	&	0.926	&	0.956	\\
\emph{Mistral-7B-OpenOrca}	 &	&	0.986	&	0.861	&	0.920	\\
\emph{Zephyr-7B-Alpha}	 &	&	0.981	&	0.910	&	0.944	\\
\emph{Zephyr-7B-Beta}	 &	&	0.982	&	0.914	&	0.950	\\
\emph{Ensemble}		 &	&	\textbf{0.988}	&	\textbf{0.931}	&	\textbf{0.962}	\\	
\hline
\end{tabular}
\caption{LLMs comparison for the information extraction task of the country name over IDB.}\label{country}
\end{table}

\noindent
Each IE task was assessed as a binary classification problem. In this context, the negative class means that no information (``None'') is linked to the IDB data sample, while the positive class indicates that the IDB is labelled with a specific epidemiological information. The models were tested on their ability to recognize such positive and negative classes by considering the 
%
%
widely adopted \emph{Precision}, \emph{Recall}, and $F_1$ \emph{score} metrics~\cite{DBLP:books/sp/CRP2019} for each class: 
%
\begin{equation} \label{eq:precision}
 Precision = \frac{TP}{TP+FP},
\end{equation}
\begin{equation} \label{eq:recall}
 Recall = \frac{TP}{TP+FN},
\end{equation}
\begin{equation} \label{eq:f1score}
 F_1 = \frac{TP}{TP+(FP+FN)/2} = \frac{2}{\frac{1}{Precision}+\frac{1}{Recall}},
\end{equation}
where $TP$, $FP$, $FN$, and $TN$ are, respectively, the number of True Positives, the number of False Positives, the number of False Negatives, and the number of True Negatives~\cite{DBLP:books/sp/CRP2019}.
In words, the \emph{Precision} (or \emph{Positive Predictive Value}) is the proportion of the class instances with a positive classification which are correctly detected. 
The \emph{Recall} (or \emph{Sensitivity}) is the proportion of the positive class instances that are correctly detected as such by the classifier. The $F_1$ \emph{score} is the harmonic mean of precision and recall: whereas the regular mean treats all values equally, the harmonic one gives much weight to low values, and for this reason is a metric preferred to the classical accuracy in an imbalanced problem setting. We obtain a large $F_1$ for the positive class only if its precision and recall are high.

\noindent
For the computational analysis, we considered the LLMs used in \emph{Ensemble}, namely 
\emph{Mistral-7B-OpenOrca}, \emph{Zephyr-7B-Beta}, and \emph{Meta-Llama-3-70B-Instruct}, as well as the 
open-source LLMs versions adopted in \cite{Consoli2024241ICICT}, namely \emph{Pythia-12B} \cite{Biderman20232397}, 
\emph{MPT-30B-Chat} \cite{MosaicML2023IntroducingMPT}, 
and \emph{Meta-Llama-2-70B-Chat} \cite{touvron2023llama}. 
We further considered in the analysis the commercial OpenAI's GPT models family \cite{openaiGPTModels2024}, 
namely \emph{GPT-3.5-Turbo-16k}, a.k.a ChatGPT and \emph{GPT-4-32k}, along with a \emph{GPT-4-FewShots} variant by passing three examples (Three-Shots) of epidemiological information extraction in the context to the \emph{GPT-4-32k} prompt, such that to try to specialize the model 
to handle the specific epidemiological IE task. In-context learning (ICL) \cite{Brown2020} refers to a machine learning approach where a model learns from the context. Contrary to supervised learning which requires a training phase that utilizes backward gradients to update model parameters, ICL bypasses parameter updates and directly carries out predictions on the pretrained language models. The expectation is that the model will identify the pattern concealed in the examples and make the correct prediction accordingly \cite{dong2023survey}. We leveraged the ICL approach on the \emph{GPT-4-32k} model, leading to the \emph{GPT-4-FewShots} variant to allow for a more efficient and accurate way to extract and process information on epidemics. Please note that ICL was only applied to \emph{GPT-4-32k} due its large context capability, which was not amenable in the other studied LLMs which are characterised by a way lower context.

\noindent
Please note that the LLMs employed in this study were used through the GPT@JRC initiative of the Joint Research Centre (JRC) of the European Commission, which enables JRC staff to explore the potential uses of AI pre-trained LLMs. The initiative, which is part of a broader study on the new technology's applications within the European Commission, is a central hub offering secure access to various AI models. 
GPT@JRC is hosted at the JRC datacentre and supports both open-source AI models, deployed on-premises at JRC Big Data Analytics Platform \cite{Soille201830}, 
and commercial OpenAI's GPT models running in the European Cloud under a Commission contract with 
an opt-out clause on prompt analysis by third parties.

\begin{table}[!b]
\centering
\begin{tabular}{|rl|r|r|r|}
\hline
 & & \multicolumn{1}{c|}{\textbf{Precision}} & \multicolumn{1}{c|}{\textbf{Recall}} & \multicolumn{1}{c|}{\textbf{F$_{1}$ score}} \\
\hline
\emph{GPT-3.5-Turbo-16k}	 &	&	0.699	&	0.455	&	0.551	\\
\emph{GPT-4-32k}			 &	&	0.673	&	0.589	&	0.629	\\
\emph{GPT-4-FewShots}		 &	&	0.733	&	0.589	&	0.653	\\
\emph{Pythia-12B}			 &	&	0.365	&	0.277	&	0.315	\\
\emph{MPT-30B-Chat}		 &	&	0.619	&	0.464	&	0.531	\\
\emph{Meta-Llama-2-70B-Chat}	 &	&	0.561	&	0.536	&	0.548	\\
\emph{Meta-Llama-3-70B-Instruct} &	&	0.722	&	0.590	&	0.652	\\
\emph{Mistral-7B-OpenOrca}	 &	&	0.67	&	0.527	&	0.590	\\
\emph{Zephyr-7B-Alpha}	 &	&	0.615	&	0.571	&	0.593	\\
\emph{Zephyr-7B-Beta}	 &	&	0.650	&	0.585	&	0.612	\\
\emph{Ensemble}		 &	&	\textbf{0.788}	&	\textbf{0.591}	&	\textbf{0.658}	\\
\hline
\end{tabular}
\caption{LLMs comparison for the information extraction task of the number of confirmed cases over IDB.}\label{cases}
\end{table}

\begin{table}[!b]
\centering
\begin{tabular}{|rl|r|r|r|}
\hline
 & & \multicolumn{1}{c|}{\textbf{Precision}} & \multicolumn{1}{c|}{\textbf{Recall}} & \multicolumn{1}{c|}{\textbf{F$_{1}$ score}} \\
\hline
\emph{GPT-3.5-Turbo-16k}	 	& &	0.902	&	0.639	&	0.748	\\
\emph{GPT-4-32k}			 	& &	0.913	&	0.734	&	0.814	\\
\emph{GPT-4-FewShots}		 	& &	0.904	&	0.658	&	0.762	\\
\emph{Pythia-12B}			 	& &	0.735	&	0.158	&	0.260	\\
\emph{MPT-30B-Chat}		 	& &	0.814	&	0.304	&	0.442	\\
\emph{Meta-Llama-2-70B-Chat}	 	& &	0.916	&	0.759	&	0.830	\\
\emph{Meta-Llama-3-70B-Instruct} 	& &	0.922	&	0.807	&	0.862	\\
\emph{Mistral-7B-OpenOrca}	 	& &	0.908	&	0.690	&	0.784	\\
\emph{Zephyr-7B-Alpha}	 	& &	0.920	&	0.804	&	0.858	\\
\emph{Zephyr-7B-Beta}	 	& &	0.921	&	0.810	&	0.861	\\
\emph{Ensemble}		 	& &	\textbf{0.928}	&	\textbf{0.813}	&	\textbf{0.869}	\\
\hline
\end{tabular}
\caption{LLMs comparison for the information extraction task of the disease outbreak date over IDB.}\label{date}
\end{table}

\noindent
Tables \ref{epidemicname}-\ref{date} report our computational results on the comparison of the LLMs for the extraction of, respectively, the disease outbreak names, the countries where the outbreak occurred, the related event date, and the number of confirmed cases associated to the specific disease outbreak. 
Our analysis reveals that the \emph{Pythia-12B} and \emph{MPT-30B-Chat} models exhibit lower performance compared to their counterparts. These models, while still valuable, may not be the optimal choice for applications requiring the highest level of accuracy in epidemiological IE tasks.

Commercial LLMs, particularly \emph{GPT-4-32k}, demonstrated superior performance, aligning with expectations given their advanced architecture and extensive training datasets. \emph{GPT-3.5-Turbo-16k} also shows a significant performance, indicating its potential utility in epidemiological IE tasks. The application of in-context learning, specifically the \emph{GPT-4-32k-FewShots} variant, showed that providing the model with a few examples can enhance its performance, although this improvement is not consistent across all tasks.

Despite the high performance of OpenAI's GPT models, their implementation is associated with significant costs and practical limitations, such as usage restrictions, which may hinder their widespread deployment on large datasets like the full DONs.

In the realm of open-source models, we observe notable advancements in performance with newer iterations. The \emph{Meta-Llama-3-70B-Instruct} model outperforms its predecessor, \emph{Meta-Llama-2-70B-Chat}, and similarly, \emph{Zephyr-7B-Beta} shows improvements over \emph{Zephyr-7B-Alpha}, as we were expecting. These enhancements underscore the rapid progress within the field of open-source LLMs.

Among the open-source LLMs, \emph{Meta-Llama-3-70B-Instruct}, \emph{Mistral-7B-OpenOrca}, and \emph{Zephyr-7B-Beta} stand out for their high performance, rivaling and occasionally surpassing that of the commercial GPT models. It's important to note that in our computational experience the \emph{Meta-Llama-3-70B-Instruct} model exhibited slower computational times due to its large number of parameters (70 billions), which may impact its practicality in time-sensitive applications.

\begin{figure}[!b]
\centering
\adjustbox{cfbox=gray 1pt}{
\includegraphics[width=0.8\linewidth]{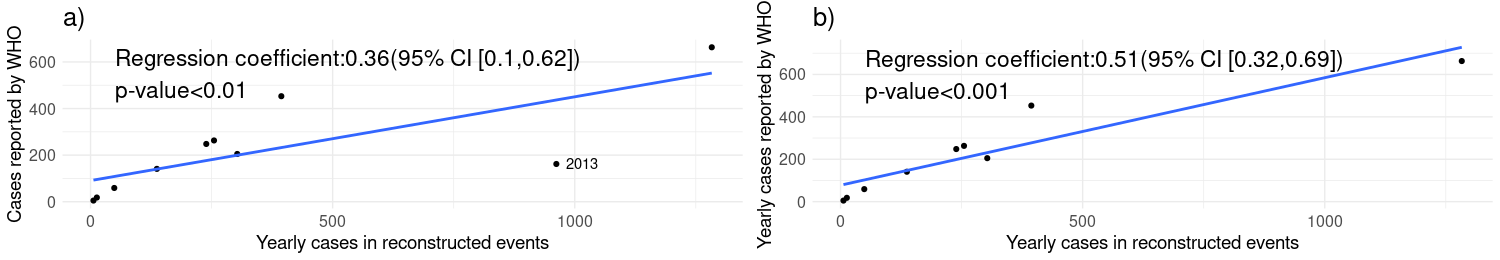}
}
\caption{WHO yearly number of cases vs the number of cases from the reconstructed events, together with regression line: including year 2013 (Panel a) and without year 2013 (Panel b). Since 2013 was the epidemic's onset, the inclusion of potential and confirmed cases by WHO DONS could lead to a lower correlation coefficient due to overestimated reconstructed cases. Correlation and p-values are reported in the top right corner of each panel.}
\label{fig_corr}
\end{figure}

The \emph{Ensemble} approach, which combines the strengths of \emph{Mistral-7B-OpenOrca}, \emph{Meta-Llama-3-70B-Instruct}, and \emph{Zephyr-7B-Beta}, emerges as the most effective strategy (see results in bold in the tables). This method not only achieves the highest F$_{1}$ scores across all four tasks but also addresses the limitations of individual models. The ensemble's robust performance across diverse IE tasks suggests its suitability for full-scale deployment on the entire DONs dataset, providing a reliable and efficient solution for extracting structured epidemiological information, and supporting the technical quality of the extracted eKG dataset. The ensemble approach leverages the collective capabilities of high-performing open-source LLMs to deliver an optimal balance of accuracy, speed, and adaptability. This method's superior performance, as evidenced by the results in the tables, supports the reliability of the produced dataset and its large potential 
for epidemic outbreak modeling and surveillance systems.

%



\noindent
From a more qualitative point of view, 
we compared the yearly number of MERS-Cov cases in Saudi Arabia from the reconstructed events with the number of confirmed cases reported by the WHO~\cite{WHO_mers}. 
Figure \ref{fig_corr} shows the scatter plots of WHO yearly number of cases vs the number of cases from the reconstructed events, together with the best-fit regression line, the 95\% CI of the regression coefficient and the corresponding p-value. Panel a shows the results for all data, while Panel b depicts the results obtained by excluding year 2013. As it can be observed, the regression coefficient between the reconstructed events and the WHO data is significant (Panel a: 95\% CI  of regression coefficient $[0.10,0.62]$, p-value$<0.01$) and increases considerably (Panel b: 95\% CI  of regression coefficient $[0.30,0.69]$, p-value$<0.001$) when data from year 2013 is not considered. 
Considering 2013 was the onset of the epidemic, it is plausible that the WHO DONS might have included potential cases along with confirmed ones. This could result in an overestimation of the reconstructed cases compared to the confirmed ones, resulting in a lower regression coefficient if data from year 2013 are taken into consideration.

\section*{Usage Notes}
%
%

In situations where policy-makers have to act fast to reduce economic and societal downturns, real-time forecasting of a epidemic's intensity becomes essential \cite{kissler2020}. As a result, researchers and policy-makers are increasingly looking for alternative data sets that can compensate for the lack of timeliness in official statistics \cite{paolotti2014}. These opportunities and challenges in infectious disease epidemiology \cite{Lewnard2019873} are inspiring some of the research activities of the Joint Research Centre (JRC) \cite{JRCEC} 
of the European Commission (EC), linking to other ongoing relevant initiatives on the subject \cite{whoeios2024}. 
Ongoing work aims to track outbreaks in the European Union (EU) using data sets that are considered unconventional in classic epidemiological modeling, and the development of eKG falls in this context. 
We enable 
the interested community 
to use the extracted data under Creative Commons Attribution 4.0 International (CC BY 4.0) license \cite{CCBY4}. In particular, the epidemiological information extracted from WHO DONs are stored in the RDF/XML named-graph available at: \href{https://jeodpp.jrc.ec.europa.eu/ftp/jrc-opendata/ETOHA/ETOHA-OPEN/epidemicIE-DONs.rdf}{Epidemiological knowledge graph (eKG) from WHO DONs using LLMs - RDF}, also available in Turtle format \cite{RDFturtle2014} 
at: \href{https://jeodpp.jrc.ec.europa.eu/ftp/jrc-opendata/ETOHA/ETOHA-OPEN/epidemicIE-DONs.ttl}{Epidemiological knowledge graph (eKG) from WHO DONs using LLMs - TTL}, with raw CSV extractions in: \href{https://jeodpp.jrc.ec.europa.eu/ftp/jrc-opendata/ETOHA/corpus_processed/SUMMARIZED/OutputAnnotatedTexts-LLMs-ENSEMBLE_whoDons.csv}{Epidemiological IE from WHO DONs using LLMs - Raw Extractions}.

\begin{figure}[!b]
\centering
\adjustbox{cfbox=gray 1pt}{
\includegraphics[width=0.75\linewidth]{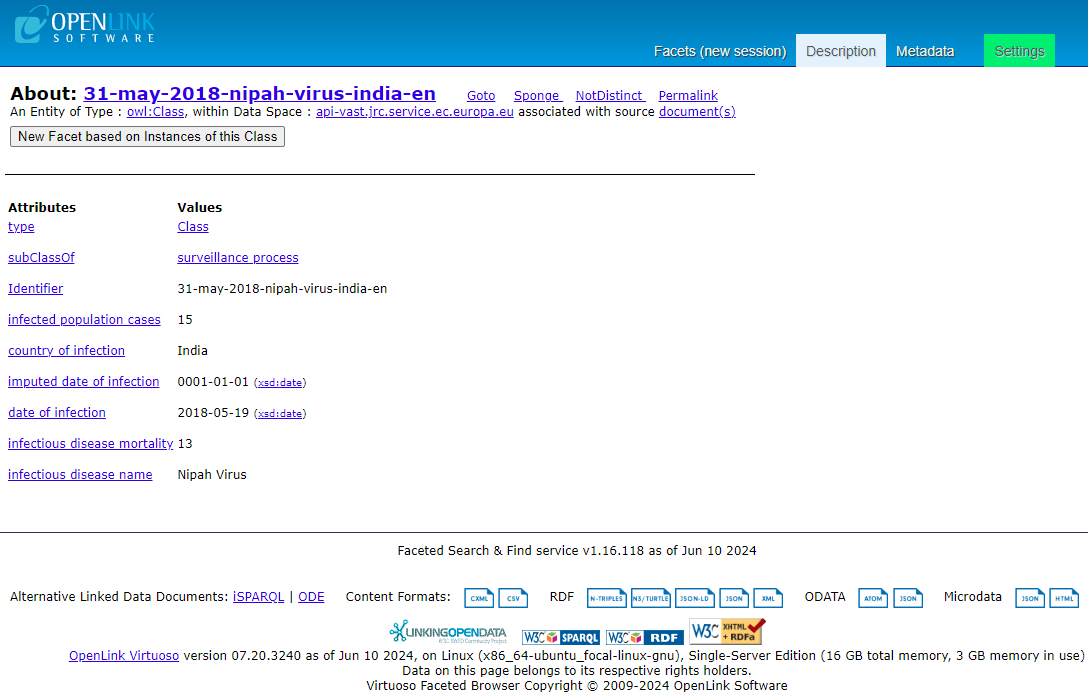}
}
\caption{Faceted Browser showing the extracted information for \href{https://www.who.int/emergencies/disease-outbreak-news/item/31-may-2018-nipah-virus-india-en}{Nipah virus - India}.}
\label{fig:faceted-nipah}
\end{figure}

\noindent
The eKG dataset can be loaded within any triplestore supporting it. 
In particular, our eKG can be accessed for exploration in the live OpenLink Virtuoso triplestore, available at \url{https://api-vast.jrc.service.ec.europa.eu/sparql/} \cite{sparqlEndpoitEPIKG}. 
The knowledge graph is interrogable via SPARQL queries 
by editing the text area available in the Virtuoso interface for the SPARQL query language, which is the standard reference language and a W3C \cite{Griset2011353W3C} 
recommendation for querying RDF/OWL data. The SPARQL endpoint is also accessible as a dedicated REST web service \cite{sparqlEndpoitEPIKG}. It requires as input a user-defined SPARQL query and produces as output the query result in one of the following formats: \emph{text/html}, \emph{text/rdf +n3}, \emph{application/xml}, \emph{application/json}, or \emph{application/rdf+xml}. \\
Queries such as \texttt{CONSTRUCT}, \texttt{ASK}, \texttt{DESCRIBE}, and \texttt{SELECT} can be executed to retrieve data from eKG. A \texttt{CONSTRUCT} query yields an RDF graph created by replacing variables within the query. An \texttt{ASK} query produces a boolean value, either true or false, signifying if the query pattern has any matches. A \texttt{SELECT} query outputs the values of variables that are successfully matched within the query pattern. Typically, this kind of query is structured into three main sections:\\
\indent 1. \texttt{PREFIX} is used to define shorthand prefixes for URIs in the query.\\
\indent 2. \texttt{SELECT} specifies which variables will be included in the outcome of the query.\\
\indent 3. \texttt{FROM} indicates to which named-graph is requested to restrict the query.\\
\indent 4. \texttt{WHERE} sets out the fundamental graph pattern that the data must correspond to.


\begin{figure}[!t]
\centering
\adjustbox{cfbox=gray 1pt}{
\includegraphics[width=0.75\linewidth]{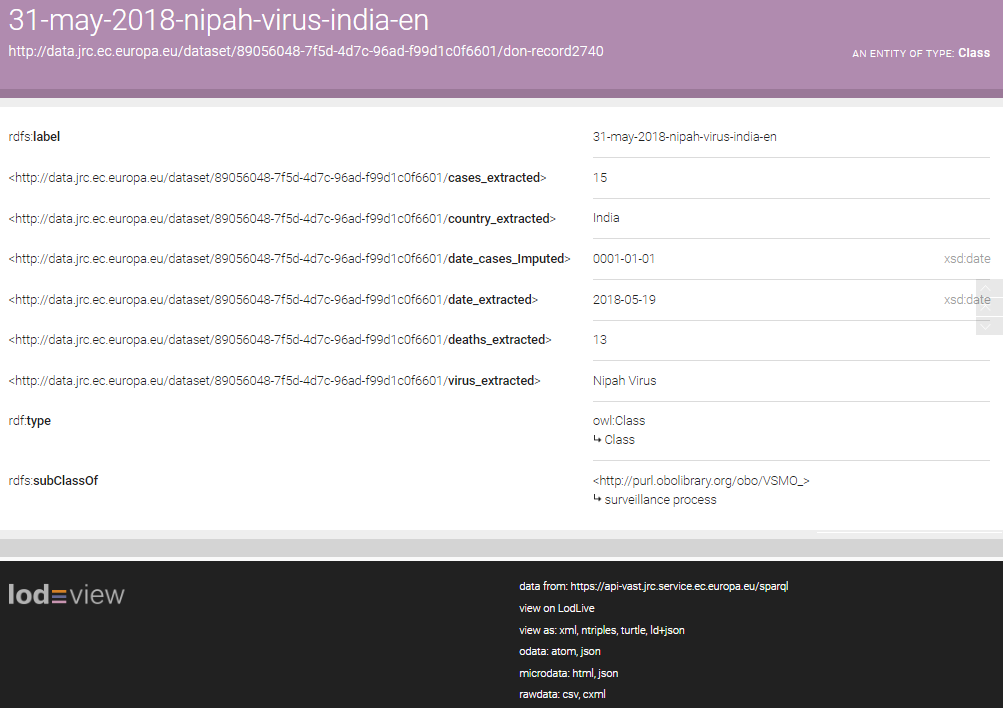}
}
\caption{Content negotiation of the \href{https://www.who.int/emergencies/disease-outbreak-news/item/31-may-2018-nipah-virus-india-en}{Nipah virus - India} report by means of LodView.}
\label{fig:lodview-nipah}
\end{figure}

\begin{figure}[!th]
\centering
\adjustbox{cfbox=gray 1pt}{
\includegraphics[width=0.75\linewidth]{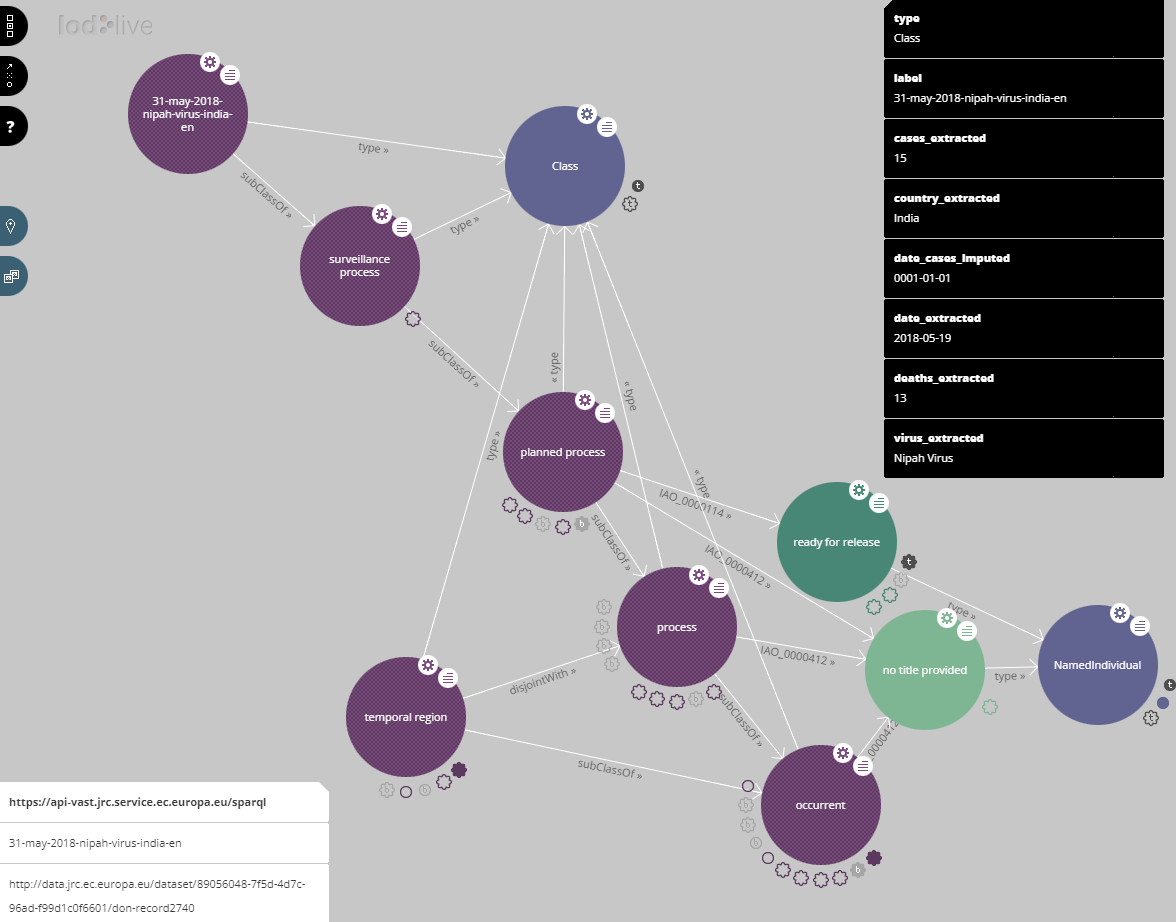}
}
\caption{Graph visualization of \href{https://www.who.int/emergencies/disease-outbreak-news/item/31-may-2018-nipah-virus-india-en}{Nipah virus - India} in LodLive.}
\label{fig:lodlive-nipah}
\end{figure}

\noindent
Consider the \href{https://www.who.int/emergencies/disease-outbreak-news/item/31-may-2018-nipah-virus-india-en}{Nipah virus - India} example, having id: \texttt{don-record2740}. Recalling the eKG namespace for the data resources being \texttt{http://data.jrc.ec.europa.eu/dataset/89056048-7f5d-4d7c-96ad-f99d1c0f6601}, 
the RDF information related to this individual can be obtained from the endpoint with the SPARQL query:
\begin{mybox}
\texttt{PREFIX eKG: \\<http://data.jrc.ec.europa.eu/dataset/89056048-7f5d-4d7c-96ad-f99d1c0f6601/>}\\[2mm]
\href{https://api-vast.jrc.service.ec.europa.eu/sparql?default-graph-uri=&query=SELECT+*+where+%7B+%3Chttp%3A%2F%2Fdata.jrc.ec.europa.eu%2Fdataset%2F89056048-7f5d-4d7c-96ad-f99d1c0f6601%2Fdon-record2740%3E+%3Fp+%3Fo.%7D+&format=text%2Fhtml}{
\texttt{SELECT *\\
FROM <eKG> \\
WHERE \{eKG:don-record2740 ?p ?o\}}}
\end{mybox}

\noindent
In simple words, this query translates into the selection of all the data triples having properties ?p and objects ?o, matching with the resource ``Nipah virus - India'' with \texttt{don-record2740} id specified as subject. The \texttt{don-record2740} can be visualized with the Faceted Browser as in \url{https://api-vast.jrc.service.ec.europa.eu/describe//?url=http://data.jrc.ec.europa.eu/dataset/89056048-7f5d-4d7c-96ad-f99d1c0f6601/don-record2740}, and also illustrated in Figure \ref{fig:faceted-nipah}, where users can explore the RDF information of the \href{https://www.who.int/emergencies/disease-outbreak-news/item/31-may-2018-nipah-virus-india-en}{Nipah virus - India} report in a structured and intuitive way, providing 
a classification of the data in various dimensions. 
In Figure \ref{fig:lodview-nipah}, it is also reported its LodView representation 
where one can exploit and navigate the DON entity, seeing for example the related outbreak information, the report categorisation as \texttt{IDO:surveillance\_process}
and \texttt{OWL:Class}, 
and downloading the raw data in one of the available formats, e.g. \href{https://lodview.it/lodview/?IRI=http://data.jrc.ec.europa.eu/dataset/89056048-7f5d-4d7c-96ad-f99d1c0f6601/don-record2740&sparql=http%3A%2F%2Fpublications.europa.eu%2Fwebapi%2Frdf%2Fsparql&prefix=http%3A%2F%2Fdata.jrc.ec.europa.eu%2Fdataset%2F&output=application/rdf+xml}{rdf+xml} or \href{https://lodview.it/lodview/?IRI=http://data.jrc.ec.europa.eu/dataset/89056048-7f5d-4d7c-96ad-f99d1c0f6601/don-record2740&sparql=http%3A%2F%2Fpublications.europa.eu%2Fwebapi%2Frdf%2Fsparql&prefix=http%3A%2F%2Fdata.jrc.ec.europa.eu%2Fdataset%2F&output=application/ld+json}{ld+json}, among others. 
Finally, Figure \ref{fig:lodlive-nipah} shows the LodLive graph representation of the RDF data of the \texttt{don-record2740} report relatively to the other outbreak events present in eKG. 
For instance this allows the user to expand automatically the relations of a selected resource, calculate inverse and \texttt{owl:sameAs} relations, store images during the navigation, and eventually geo-localize the browsed data as points in a map.


In the following boxes, we showcase some further examples of queries for developers and researchers that can be run on eKG. 
These queries are indented to assist the scientific community 
on the use of the extracted knowledge graph. To avoid repetitions, consider all the following SPARQL queries to be prefaced with the following standard headers:
\begin{mybox}
\texttt{PREFIX eKG: \\<http://data.jrc.ec.europa.eu/dataset/89056048-7f5d-4d7c-96ad-f99d1c0f6601/>\\
PREFIX rdf: <http://www.w3.org/1999/02/22-rdf-syntax-ns\#>\\
PREFIX rdfs: <http://www.w3.org/2000/01/rdf-schema\#>\\
PREFIX dcterm: <http://purl.org/dc/terms/>\\
PREFIX dc: <http://purl.org/dc/elements/1.1/> \\
PREFIX owl: <http://www.w3.org/2002/07/owl\#>\\
PREFIX xml: <http://www.w3.org/XML/1998/namespace>\\
PREFIX xsd: <http://www.w3.org/2001/XMLSchema\#>\\
PREFIX obo: <http://purl.obolibrary.org/obo/>\\
PREFIX skos: <http://www.w3.org/2004/02/skos/core\#> }
\end{mybox}

\begin{figure}[!b]
\centering
\adjustbox{cfbox=gray 1pt}{
\includegraphics[width=1.0\linewidth]{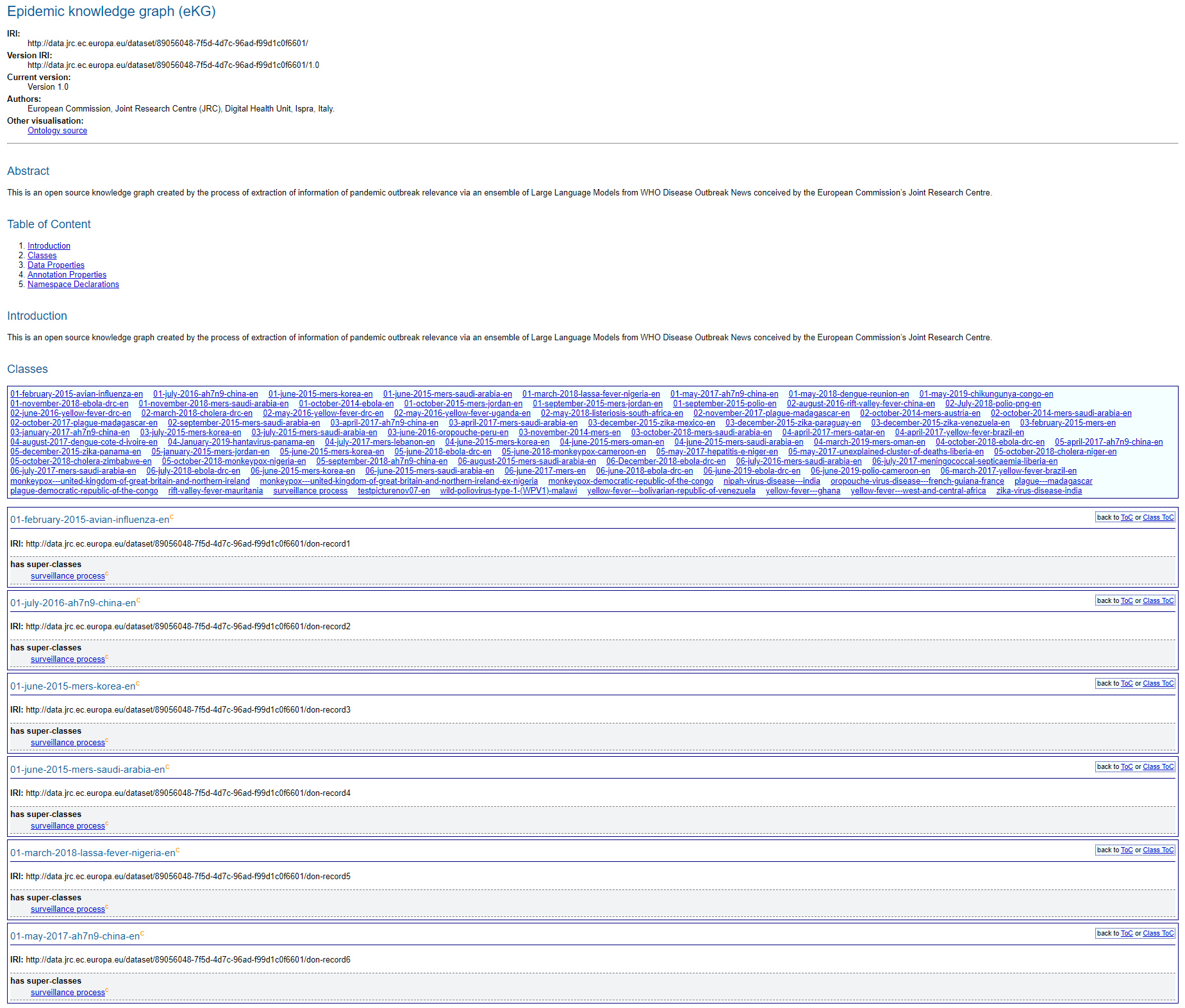}
}
\caption{An extract from the eKG representation in LODE.}
\label{fig:lode}
\end{figure}

\begin{figure}[!t]
\centering
\includegraphics[width=0.99\linewidth]{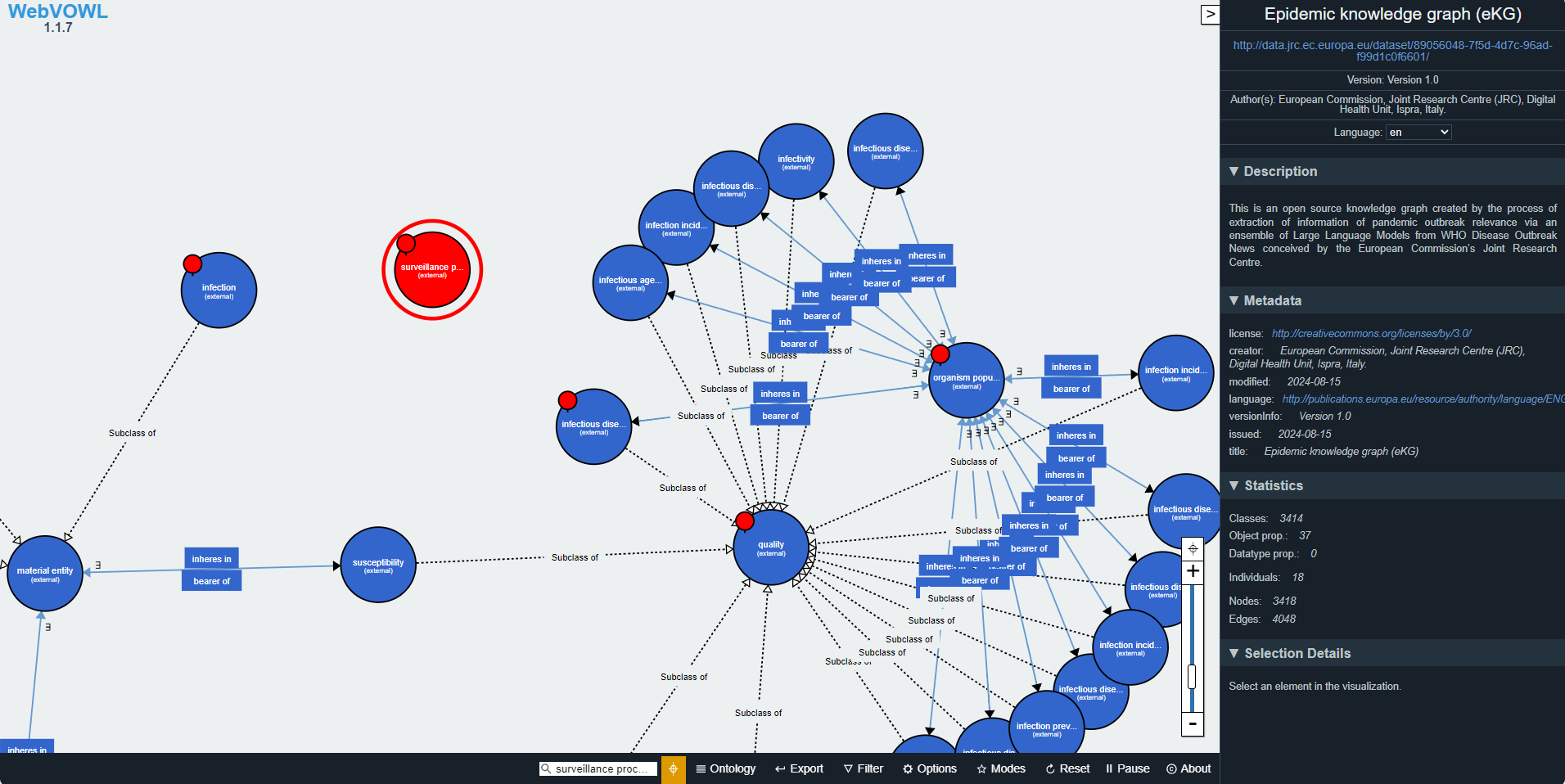}
\caption{A snapshot of the WebVOWL visualization of eKG.}
\label{fig:webvowl}
\end{figure}

\noindent
Suppose one wants to know the total number of triples of eKG, which is growing dynamically in time; this would be translated as a simple SPARQL query as follows:
\begin{mybox}
\href{https://api-vast.jrc.service.ec.europa.eu/sparql?default-graph-uri=&query=SELECT+COUNT%28*%29+from+%3CeKG%3E+WHERE+%7B+%3Fs+%3Fp+%3Fo+%7D&format=text%2Fhtml}{
\texttt{SELECT COUNT(*)\\
FROM <eKG> \\
WHERE \{?s ?p ?o}}\}
\end{mybox}

\noindent
To retrieve instead all the events related to ``Nipah virus'' outbreak:
\begin{mybox}
\href{https://api-vast.jrc.service.ec.europa.eu/sparql?default-graph-uri=&query=PREFIX+eKG%3A+%3Chttp%3A%2F%2Fdata.jrc.ec.europa.eu%2Fdataset%2F89056048-7f5d-4d7c-96ad-f99d1c0f6601%2F%3E%0D%0A%0D%0ASELECT+%3Fevent+%0D%0AFROM+%3CeKG%3E%0D%0AWHERE+%7B%0D%0A++%3Fevent+eKG%3Avirus_extracted+%3Flabel+.%0D%0A++FILTER+%28%3Flabel+%3D+%22Nipah+Virus%22%29%0D%0A%7D+&format=text%2Fhtml}{
\texttt{SELECT ?event \\
FROM <eKG> \\
WHERE \{?event eKG:virus\_extracted ?label . \\
\hspace*{15mm}FILTER (?label = ``Nipah Virus'')\}}}
\end{mybox}

\noindent
The same but with a more flexible regular expression in the SPARQL query:
\begin{mybox}
\href{https://api-vast.jrc.service.ec.europa.eu/sparql?default-graph-uri=&query=PREFIX+eKG%3A+%3Chttp%3A%2F%2Fdata.jrc.ec.europa.eu%2Fdataset%2F89056048-7f5d-4d7c-96ad-f99d1c0f6601%2F%3E%0D%0A%0D%0ASELECT+%3Fevent+%0D%0AFROM+%3CeKG%3E%0D%0AWHERE+%7B%0D%0A++%3Fevent+eKG%3Avirus_extracted+%3Flabel+.%0D%0AFILTER+regex%28str%28%3Flabel%29%2C+%22nipah%22%2C+%22i%22%29%0D%0A%7D+&format=text%2Fhtml}{
\texttt{SELECT ?event \\
FROM <eKG> \\
WHERE \{?event eKG:virus\_extracted ?label . \\
\hspace*{15mm}FILTER regex(str(?label), ``nipah'', ``i'')\}}}
\end{mybox}

\noindent
The following SPARQL query finds all the disease outbreaks that occurred in ``Italy'':
\begin{mybox}
\href{https://api-vast.jrc.service.ec.europa.eu/sparql?default-graph-uri=&query=PREFIX+eKG%3A+%3Chttp%3A%2F%2Fdata.jrc.ec.europa.eu%2Fdataset%2F89056048-7f5d-4d7c-96ad-f99d1c0f6601%2F%3E%0D%0A%0D%0ASELECT+%3Fevent+%0D%0AFROM+%3CeKG%3E%0D%0AWHERE+%7B%0D%0A++%3Fevent+eKG%3Acountry_extracted+%3Flabel+.%0D%0A+FILTER+%28%3Flabel+%3D+%22Italy%22%29%0D%0A%7D+&format=text%2Fhtml}{
\texttt{SELECT ?event \\
FROM <eKG> \\
WHERE \{?event eKG:country\_extracted ?label . \\
\hspace*{15mm}FILTER (?label = ``Italy'')\}}}
\end{mybox}

\noindent
Lastly, the following SPARQL query retrieves all the disease outbreaks occurring in ``Italy'' in the year ``2017'', specifying the name of the disease and the number of confirmed cases:
\begin{mybox}
\href{https://api-vast.jrc.service.ec.europa.eu/sparql?default-graph-uri=&query=PREFIX+eKG%3A+%3Chttp%3A%2F%2Fdata.jrc.ec.europa.eu%2Fdataset%2F89056048-7f5d-4d7c-96ad-f99d1c0f6601%2F%3E%0D%0A%0D%0ASELECT+%3Fevent++%3Foutbreak+%3Fcases%0D%0AFROM+%3CeKG%3E%0D%0AWHERE+%7B%0D%0A++%3Fevent+eKG%3Acountry_extracted+%3Flabel+.%0D%0A%3Fevent+eKG%3Adate_extracted+%3Fdate+.%0D%0A++FILTER+%28%3Flabel+%3D+%22Italy%22%29+.+%0D%0A++FILTER+%28year%28%3Fdate%29+%3D+2017%29+.+%0D%0A%3Fevent+eKG%3Avirus_extracted+%3Foutbreak+.%0D%0A%3Fevent+eKG%3Acases_extracted+%3Fcases+.%0D%0A%7D+&format=text%2Fhtml}{
\texttt{SELECT ?event ?outbreak ?cases \\
FROM <eKG> \\
WHERE \{?event eKG:country\_extracted ?label . \\
?event eKG:date\_extracted ?date . \\
\hspace*{15mm}FILTER (?label = ``Italy'') . \\
\hspace*{15mm}FILTER (year(?date) = 2017) . \\
\hspace*{15mm}?event eKG:virus\_extracted ?outbreak . \\
\hspace*{15mm}?event eKG:cases\_extracted ?cases . \}}}
\end{mybox}

\noindent
These queries represent only some basic examples and there might be various other use cases that can be addressed. For more complex queries, we refer the reader to various tutorials available on the web, 
and to technical articles in the literature \cite{SPARQLPerez2009,SPARQLSchmidt20104} for a more intense deep dive into the SPARQL query language.

\noindent
The knowledge graph of epidemiological information can be also browsed as a whole in human-readable format by means of \emph{Live OWL Documentation Environment (LODE)} facilitating easy exploration by users. LODE provides an easy-to-use interface for viewing eKG's general axioms, namespace declarations, named individuals, classes, annotations, and object properties within intuitive HTML pages: \url{http://wit.istc.cnr.it/arco/lode/extract?url=https://jeodpp.jrc.ec.europa.eu/ftp/jrc-opendata/ETOHA/ETOHA-OPEN/epidemicIE-DONs.rdf}. Figure \ref{fig:lode} shows an extract from the eKG representation in LODE. The knowledge graph can also be accessed through a holistic, force-directed graph layout using \emph{WebVOWL}: 
\url{https://service.tib.eu/webvowl/\#iri=https://jeodpp.jrc.ec.europa.eu/ftp/jrc-opendata/ETOHA/ETOHA-OPEN/epidemicIE-DONs.rdf}, see also Figure \ref{fig:webvowl}. Interaction with these tools allows for customizable visualizations, offering a user-friendly description of the elements of eKG and enhancing the user's ability to leverage the epidemiological information within.

\section*{Code availability}
%

The codebase and documentation of the pipeline used for epidemiological information extraction from WHO DONs with the \emph{Ensemble} of LLMs and the generation of eKG can be found at: \url{https://github.com/jrcf7/epidemicIE-eKG-dons}. 
The code was written in Python 3.10.11 and we used, in particular Pandas (\url{https://pandas.pydata.org/}) for handling the raw extraction files (e.g. CSV), Numpy (\url{https://numpy.org/}) for data preprocessing and feature extraction, Scikit-learn (\url{https://scikit-learn.org/}) for common machine learning routines, WordNet (\url{https://wordnet.princeton.edu/}) as lexical database for the English language, available via NLTK (\url{https://www.nltk.org/}), assisting in word sense disambiguation and synonymy checking, Sentence Transformers SBERT (Sentence-BERT: \url{https://sbert.net/}) to derive semantically meaningful sentence embeddings for semantic similarity at the sentence level, the PyTorch (\url{https://pytorch.org/}) framework for handling the LLMs workflow, and LangChain (\url{https://www.langchain.com/}) for data processing and integration with the LLMs through modular components and pre-built libraries. We used the GPT@JRC initiative of the Joint Research Centre (JRC) of the European Commission to have secure access to the various pre-trained LLMs employed in this study. Finally, the \textit{Prot\'eg\'e} editor (\url{http://protege.stanford.edu}) was employed to manage, adjust and maintain the knowledge graph structure.

%


\bibliography{biblio}


\section*{Acknowledgements} 
We would like to thank the colleagues of the Digital Health Unit (JRC.F7) at the Joint Research Centre of the European Commission for helpful guidance and support. 
The views expressed are purely those of the authors and may not in any circumstance be regarded as stating an official position of the European Commission. 
This research did not receive any specific grant from funding agencies in the public, commercial, or nonprofit sectors. 
We would like to acknowledge the GPT@JRC initiative for providing access to the LLMs used in this study. 
We extend our gratitude also to the JRC Big Data Analytics Platform for 
providing secure access to various open-source AI models. 
Finally, we would like to thank the colleagues contributing to the WHO Epidemic Intelligence from Open Sources (EIOS) initiative for the helpful suggestions and support during the development of this work.

\section*{Author contributions statement}


S.C. conceptualized the work, developed the pipeline, and created the knowledge graph release. 
P.V.M., N.I.S. and M.C. conceptualized and supervised the work.
L.B. and P.C. edited the original manuscript and analysed the results.
N.S. helped in developing the software.
I.B., L.S. and L.O. conceived the data collection and curated the dataset.
All authors reviewed the manuscript.

\section*{Competing interests} 
%

The authors declare no competing interests.

\end{document}